\documentclass[runningheads]{llncs}
\usepackage{graphicx}
\usepackage[colorlinks=true,
            linkcolor=blue,
            citecolor=green,
            ]{hyperref}
\usepackage{tikz}
\usepackage{comment}
\usepackage{amsmath,amssymb} % define this before the line numbering.
\usepackage{color}
\usepackage{multirow}
\usepackage{orcidlink}
\usepackage{booktabs}
\usepackage[accsupp]{axessibility}  % Improves PDF readability for those with disabilities.

\begin{document}
% \linenumbers
\pagestyle{headings}
\mainmatter
\def\ECCVSubNumber{1721}  % Insert your submission number here

\title{Learning Prior Feature and Attention Enhanced Image Inpainting} % Replace with your title

%\begin{comment}
\titlerunning{ECCV-22 submission ID \ECCVSubNumber} 
\authorrunning{ECCV-22 submission ID \ECCVSubNumber} 
\author{Anonymous ECCV submission}
\institute{Paper ID \ECCVSubNumber}
%\end{comment}
%******************

% \begin{comment}
\titlerunning{Learning Prior Feature and Attention Enhanced Image Inpainting}
% \author{Chenjie Cao\inst{1}\orcidID{0000-1111-2222-3333} \and
% Qiaole Dong\inst{2,3}\orcidID{1111-2222-3333-4444} \and
% Yanwei Fu\inst{3}\orcidID{2222--3333-4444-5555}}
\author{Chenjie Cao$^*$\orcidlink{0000-0003-3916-2843} \and
Qiaole Dong$^*$\orcidlink{0000-0002-3083-5143} \and
Yanwei Fu$^\dag$\orcidlink{0000-0002-6595-6893}}
\authorrunning{C. Cao et al.}
\institute{School of Data Science, Fudan University \\
\email{\{20110980001,qldong18,yanweifu\}@fudan.edu.cn}}
% \end{comment}
% ******************
\maketitle

\renewcommand{\thefootnote}{\fnsymbol{footnote}}
\footnotetext{$*$ Equal contributions. $\dag$ Corresponding authors.}

\begin{abstract}
Many recent inpainting works have achieved impressive results by leveraging Deep Neural Networks (DNNs) to model various prior information for image restoration.
Unfortunately, the performance of these methods is largely limited by the representation ability of vanilla Convolutional Neural Networks (CNNs) backbones.
On the other hand, Vision Transformers (ViT) with self-supervised pre-training have shown great potential for many visual recognition and object detection tasks. A natural question is whether the inpainting task can be greatly benefited from the ViT backbone?
However, it is nontrivial to directly replace the new backbones in inpainting networks, as the inpainting is an inverse problem fundamentally different from the recognition tasks. To this end, this paper incorporates the pre-training based Masked AutoEncoder (MAE) into the inpainting model, which enjoys richer informative priors to enhance the inpainting process. Moreover, we propose to use attention priors from MAE to make the inpainting model learn more long-distance dependencies between masked and unmasked regions. Sufficient ablations have been discussed about the inpainting and the self-supervised pre-training models in this paper. Besides, experiments on both Places2 and FFHQ demonstrate the effectiveness of our proposed model. Codes and pre-trained models are released in \url{https://github.com/ewrfcas/MAE-FAR}.
\end{abstract}

\section{Introduction}
\label{sec:intro}

\begin{figure}
\begin{centering}
\includegraphics[width=0.95\linewidth]{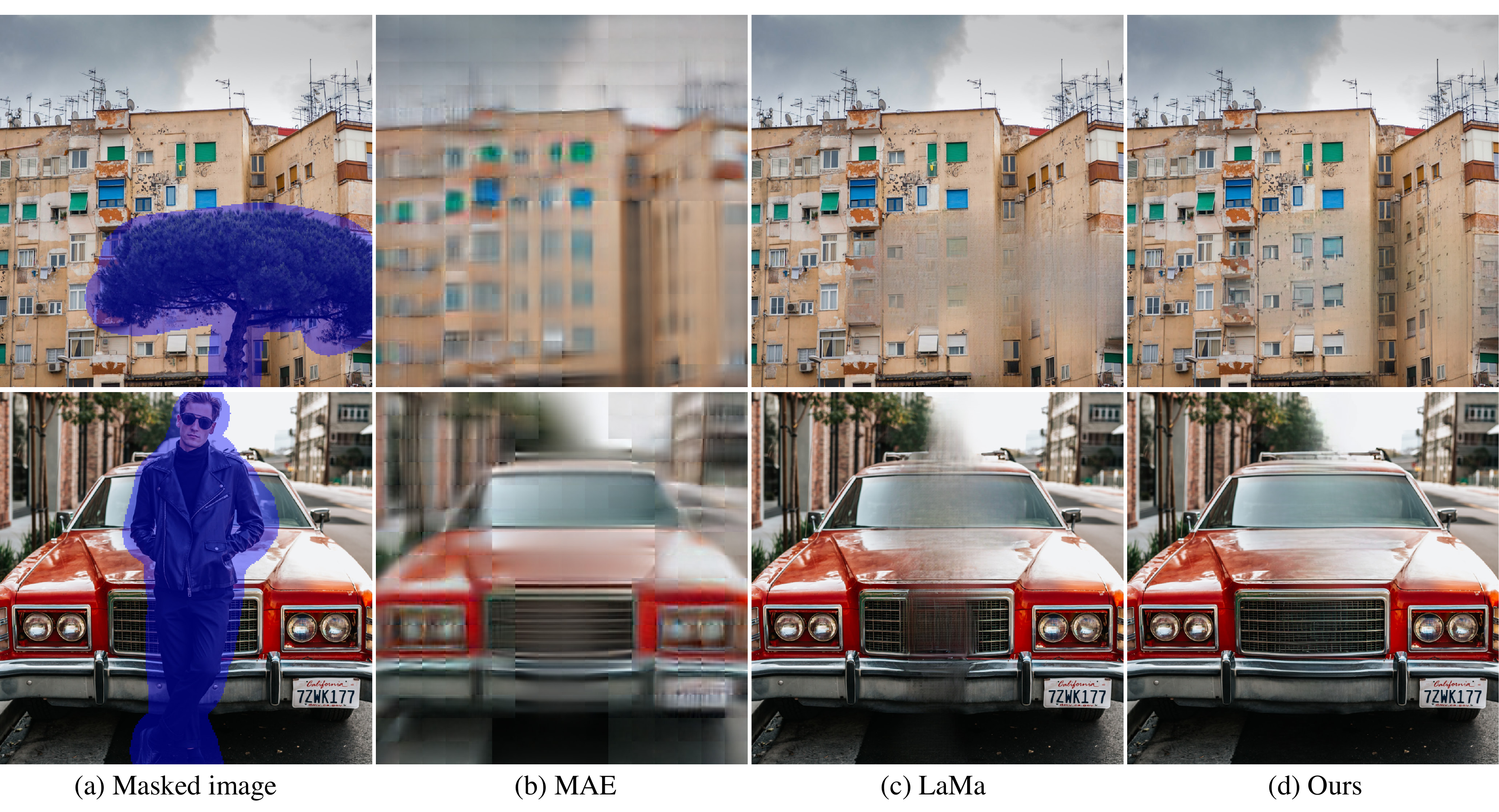}
\par\end{centering}
% \vspace{-0.15in}
\caption{The comparison of 1024$\times$1024 high-resolution image inpainting. From left to right are masked images, results of pre-trained Masked AutoEncoder (MAE)~\cite{He2021MaskedAA}, results of LaMa~\cite{suvorov2021resolution}, and results from our method.\label{fig:teaser}}
% \vspace{-0.15in}
\end{figure}

Image inpainting aims to fill missing regions of images with semantically consistent and harmoniously textured contents. It has a wide range of practical applications, including object removal~\cite{elharrouss2020image}, image editing~\cite{jo2019sc}, and so on. Conventional inpainting algorithms~\cite{Bertalmo2000ImageI,Levin2003LearningHT,Roth2005FieldsOE,Hays:2007,Criminisi2003ObjectRB} rely on visual low-level assumptions and structures to search similar patches for the reconstruction. But they fail to tackle complicated inpainting situations with the limited feature representation.

% The algorithm must generate filling content that is semantically consistent and textured harmoniously with the surrounding area in order to provide a satisfactory repair result. Many effective models in the field of image inpainting have evolved throughout the previous decades of research; yet, the problem is still far from being solved, particularly for the inpainting of high-resolution images.

% Conventional image inpainting algorithms~\cite{Bertalmo2000ImageI,Levin2003LearningHT,Roth2005FieldsOE,Hays:2007,Criminisi2003ObjectRB}, which rely on strong low-level assumptions, typically leverage some unmasked areas' manufactured low-level features (\emph{e.g.} color and texture descriptors), people's prior knowledge (\emph{e.g.} smoothness and image statistics), or even external image databases. Although they can be successful in some specific scenarios; they cannot handle complicated image scenarios, tend to output meaningless contents with inconsistent semantics, and demand extremely long runtimes for common resolution photos, which precludes their use in online image inpainting.

Deep Neural Networks (DNNs) have recently been exploited by researchers to achieve prominent improvements in image inpainting~\cite{nazeri2019edgeconnect,liao2020guidance,cao2021learning,suvorov2021resolution,wan2021highfidelity,zhao2021large}, which thanks to the great capabilities of Convolutional Neural Networks~(CNNs)~\cite{krizhevsky2012imagenet}, Generative Adversarial Networks~(GAN)~\cite{goodfellow2014generative}, and attention-based Transformers~\cite{vaswani2017attention}. However, repairing images corrupted by arbitrary masks with reasonable results is still challenging, especially in high-resolution cases. Because the inpainting model needs to understand the semantic information from masked images, which demands data-driven priors and sufficient model capacities. Furthermore, the dilemmas shown below should be solved.

\noindent \textbf{(i)} \emph{Limited capacities for modeling good priors.} 
Many pioneering works have tried to introduce prior information to inpainting models. Some works~\cite{cao2021learning,nazeri2019edgeconnect,song2018spg,yu2021wavefill,wan2021highfidelity,yu2021diverse} propose multi-stage models, which repair various auxiliary information and corrupted images sequentially to enhance the image inpainting. 
These methods learn priors in specific fields, such as structures~\cite{nazeri2019edgeconnect,cao2021learning} or semantics~\cite{song2018spg} with good visual interpretability rather than features with more informative priors.
Other methods~\cite{liao2020guidance,DBLP:journals/corr/abs-2002-04170} leverage auxiliary losses to introduce additional prior information without extending sufficient model capacities. Complex loss functions cause sophisticated hyper-parameter tuning and a more difficult inpainting training. 
Besides, some transformer based inpainting methods~\cite{wan2021highfidelity,yu2021diverse} heavily depend on low-resolution (LR) images generated by the time-consuming iterative sampling, and then upsample them with CNNs.
Lahiri \emph{et al.}~\cite{lahiri2020prior} learn global latent priors with GAN, which can only solve simple scenes with single-object.
Our method incorporates effective prior features from the transformer based representation learning~\cite{He2021MaskedAA} to enhance the inpainting, which make our method achieve superior results without overfitting the transformer results.
% (\emph{e.g.} edges, wireframes~\cite{huang2018learning}, segmentation maps, wavelets, and low-resolution images) 
% Compared with informative prior features, these decoded priors are degraded for visual interpretability and need to be re-encoded by learning another generator. Such a `decoding-encoding' process loses valuable prior information.
% Wang \emph{et al.}~\cite{wan2021highfidelity} propose to use transformer to reconstruct low-resolution (LR) images, and then upsample them with CNNs. But their results depend on the quality of LR predictions heavily rather than the prior information from the transformer. The time-consuming Gibbs iterative sampling is needed to achieve good LR results.
% Guo \emph{et al.}~\cite{guo2021image} propose to fuse structural features with a dual network during the training. But structures can not indicate semantics in various scenes, and it fails to be extended to high-resolution cases.

\noindent  \textbf{(ii)} \emph{Informative priors for high-resolution cases.} High-resolution (HR) image inpainting enjoys more practical implications with advanced electronic products and high-quality images in real-world.
Some researches devote to facilitating HR image inpainting with larger receptive fields~\cite{suvorov2021resolution,zeng2021aggregated}, attention transfer for the high-frequency residual~\cite{yi2020contextual}, and two-stage upsampling~\cite{zeng2020high}.
Unfortunately, these methods still tend to copy meaningless existing textures rather than really \emph{understand} semantics of HR masked images without directly training with costly HR data.
Our method leverages the continuous positional encoding to upsample the prior features for superior inpainting performance in HR images.

\noindent \textbf{(iii)} \emph{Missing discussions about representation learning for inpainting.} 
Recently, self-supervised pre-training language models~\cite{Radford2018ImprovingLU,Radford2019LanguageMA,Brown2020LanguageMA,Devlin2019BERTPO} have achieved great success in Natural Language Processing (NLP) fields. Such \emph{masking and predicting} idea and transformer based architectures have been also well explored in vision tasks~\cite{He2021MaskedAA,Wei2021MaskedFP,bao2021beit,zhou2021ibot}. But these vision transformers only consider representation learning for classification tasks.
To the best of our knowledge, no one has explored the application of self-supervised pre-training vision models to generative tasks, let alone image inpainting. We present comprehensive discussions about the pre-training based representation learning for image inpainting in this paper.

To address these dilemmas, we propose to guide the image inpainting with an efficient Masked AutoEncoder (MAE) pre-training model~\cite{He2021MaskedAA}, which is called as prior Feature and Attention enhanced Restoration (FAR). Specifically, an MAE is firstly pre-trained with the masked visual prediction task. We replace some random masks with large and contiguous masks to make the MAE more suitable for the downstream task. Then, features from the MAE decoder are added to the inpainting CNN for the prior guided image inpainting. Moreover, we find that attention relations of MAE among masked and unmasked regions are compatible with CNN inpainting learning. So group convolutions are used to aggregate CNN features with attention scores from MAE, which can improve the inpainting performance a lot. Furthermore, our model can be effectively extended to HR inpainting with a little finetuning of bilinear resized MAE features and the Cartesian spatial grid. Besides, we discuss some pre-training and finetuning tricks to better utilize MAE for superior inpainting performance.

% Inspired by these, we propose to guide the image inpainting model using the activation features of the pre-trained vision model, which can also be regarded as a special kind of auxiliary information with a larger amount of information than low-level image features. Specifically, we employ MAE as our pre-trained vision model, and we extract the decoder's features to guide the following image inpainting model. To give positional information, we concat a Cartesian spatial grid (CSG)~\cite{DBLP:journals/corr/JaderbergSZK15} to the decoder features. In addition, we adopt the model framework of ZITS and replace the structural feature encoder (SFE) with a simple 4-layer deconvolution layer to lift the resolution of the features, and utilize ZeroRA to inject the features into the subsequent inpainting model. Further, we also investigate the impact of MAE's loss function and masking approach on the image inpainting model's performance. Finally, our method can simply be extended to high resolution; all we need to do is interpolate the features using bilinear interpolation and then fine-tune the 
% model on high-resolution images.

We highlight our contributions as follows. 
(1) We propose to learn image priors from pre-trained MAE features, which contain informative high-level knowledge and strengthen the inpainting model.
(2) We propose to aggregate the inpainting CNN feature with attention scores from MAE to improve the performance. 
(3) Several pre-training and finetuning tricks are exploited to make our FAR learn better prior features and attentions from MAE. 
(4) Our method can be simply extended to HR cases and achieve state-of-the-art results.
Extensive experiments on Places2~\cite{zhou2017places} and FFHQ~\cite{8953766} reveal that our proposed model performs better than other competitors.

\section{Related work}

\noindent \textbf{Image Inpainting with Auxiliaries}.
Inpainting with auxiliaries has demonstrated success in many previous works. Various priors have been leveraged to enhance the inpainting, such as edges~\cite{nazeri2019edgeconnect,guo2021image}, lines~\cite{cao2021learning}, gradients~\cite{DBLP:journals/corr/abs-2002-04170}, segmentations~\cite{song2018spg,liao2020guidance}, low-resolution images~\cite{wan2021highfidelity,yu2021diverse}, and even latent priors~\cite{lahiri2020prior}. Specifically, these methods can be categorized into two types. One is to employ the approach of first correcting auxiliary information and then guiding image inpainting with multi-stage models~\cite{nazeri2019edgeconnect,lahiri2020prior,song2018spg,yu2021wavefill,cao2021learning,wan2021highfidelity,yu2021diverse}. Since these methods enjoy superior performance and good interpretability, priors leveraged by them are not comprehensive enough to handle the image inpainting properly.
Other methods~\cite{liao2020guidance,DBLP:journals/corr/abs-2002-04170} supervise the inpainting model with auxiliary information directly for introducing more positive priors. But capacities of these models are still stuck to tackle tough corrupted cases. Although work~\cite{guo2021image} combines both advantages mentioned above with the dual structure and texture learning, such low-level features are still insufficient to achieve results with rich semantics.
% costly with the `decoding and re-encoding' process of auxiliaries, which also lose useful features for the subsequent image inpainting.
% These auxiliary information can indeed provide meaningful structural guidance to the inpainting model, as well as some interpretability of the inpainted outcomes. However, they are still low-level image features in the sense that the amount of guidance they can provide to the model has a foreseeable upper limit. 

\noindent \textbf{High-resolution Image Inpainting}.
HR image inpainting with large mask areas is still challenging. In \cite{zeng2020high}, Yu \emph{et al.} propose an iterative inpainting method with a feedback mechanism to progressively fill holes. They further learn an upsampling network to handle HR inpainting results based on LR ones. Yi \emph{et al.}~\cite{yi2020contextual} design a contextual residual aggregation mechanism for the restoration of high-frequency residuals, which are added to LR predictions. Besides, various dilated convolutions are used in~\cite{zeng2021aggregated} to enlarge receptive fields for HR inpainting. Furthermore, Suvorov \emph{et al.}~\cite{suvorov2021resolution} leverage Fast Fourier Convolution (FFC) to learn a global receptive field in the frequency and achieve amazing HR inpainting results with periodic textures. However, these methods still suffer from copying meaningless textures without really `understanding' semantics in HR images. In contrast to previous methods, we transfer prior features from masked autoencoders to HR cases with a continuous positional encoding, which greatly improves the quality of HR inpainting results with meaningful semantic priors.
% they may fail to rebuild the holistic structures and perform poorly in weak texture. In contrast to previous methods, our model use masked autoencoders to learn image priors, and utilize the continuous positional encoding to upsample the prior features from the autoencoders for following inpainting model, which can greatly improve the quality of inpainted images.

\noindent \textbf{Masked Visual Prediction}.
\sloppy{Masked Visual Prediction (MVP) is a self-supervised task for representation learning by masking and predicting image patches. This work is originated in the masked language model~\cite{devlin2018bert} of NLP. The Vision Transformer (ViT)~\cite{DBLP:journals/corr/abs-2010-11929} has studied self-supervised pre-training by masking patches. BEiT~\cite{bao2021beit} and iBOT~\cite{zhou2021ibot} learn MVP on high-level discrete tokens and self-distillation respectively. Moreover, MAE~\cite{He2021MaskedAA} proposes an efficient transformer-based masked autoencoder for visual representation learning. MaskFeat~\cite{Wei2021MaskedFP} further studies MVP, and proposes to use HOG features~\cite{1467360} to get excellent results efficiently. These MVP pre-training models can be finetuned to achieve excellent classification results. However, few discussions about MVP are explored for image generation. To the best of our knowledge, our work firstly studies MVP-based pre-training for image inpainting.}
% pre-training task of directly regress the feature representation of masked regions. They searched at a wide range of feature representations, from pixels and manufactured features to activations of deep neural network, and discovered that simple manufactured features HOG~\cite{1467360} can produce excellent results, activations of deep neural network can even achieve the best pre-training effect in terms of classification performance.

\section{Method}

\begin{figure}
\begin{centering}
\includegraphics[width=0.99\linewidth]{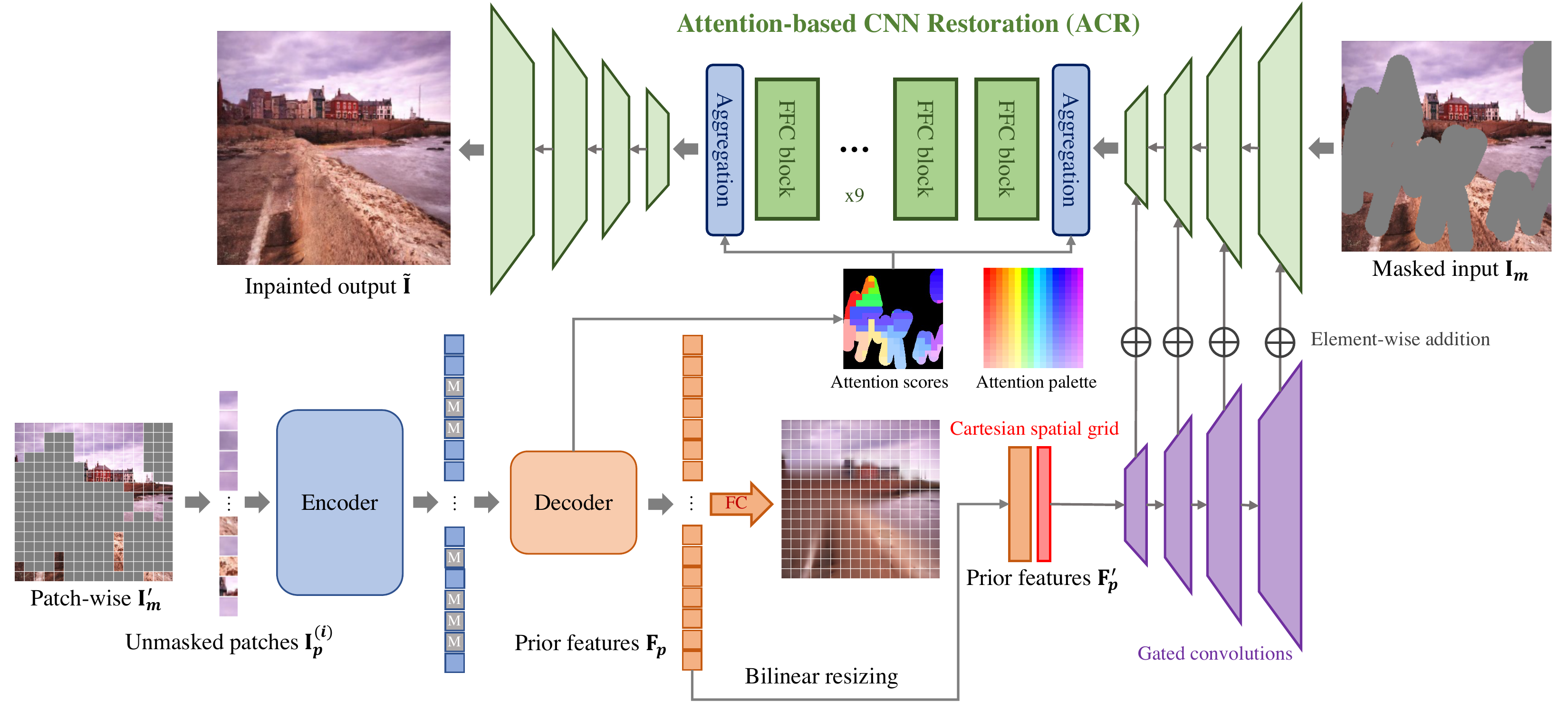}
\par\end{centering}
% \vspace{-0.1in}
\caption{The overview of our proposed FAR.}
% \vspace{-0.15in}
% We first take masked images into the MAE. Features from the decoder are resized and concatenated with the cartesian spatial grid. Then they are encoded by gated deconvolutions, and added to the encoder of ACR. Attention scores from the MAE decoder are leveraged to aggregate features from unmasked regions to masked ones for ACR.
\label{fig:overview}
\end{figure}

\noindent \textbf{Overview}.
The overall pipeline of our FAR is shown in Fig.~\ref{fig:overview}. For the given masked image $\mathbf{I}_m$, we resize it into 256$\times$256 and further enlarge the mask to patch-wise of 16$\times$16. Thus we can get the masked image $\mathbf{I}_{m}'\in\mathbb{R}^{256\times 256}$. Then the MAE is applied to encode the prior features as $\mathbf{F}_p=\mathrm{MAE}(\mathbf{I}_p^{(i)}), i\in\{1,2,...,N\}$, where $\mathbf{I}_p^{(i)}$ indicate total $N$ unmasked patches from $\mathbf{I}_{m}'$ (Sec.~\ref{sec:MAE}). Prior features are resized to $1/8$ of the original image size and concatenated with the Cartesian spatial grid as $\mathbf{F}_p'$. After that, $\mathbf{F}_p'$ is encoded by gated convolutions and added to the encoder of Attention-based CNN Restoration (ACR), which is used to restore the original masked image $\mathbf{I}_m$ (Sec.~\ref{sec:upsample}). Moreover, we leverage mean attention scores from the MAE decoder to aggregate unmasked features to masked regions in ACR and achieve the final inpainted image $\tilde{\mathbf{I}}$ (Sec.~\ref{sec:ACR}).

In this section, to better discuss the influence of pre-trained MAE on the inpainting model, we provide ablation studies on the subset of Places2~\cite{zhou2017places} with 5 scenes (about 25,000 training images, and 500 validation images, detailed in the Appendix). All methods are trained with 150k steps in 256$\times$256. Although our MAE is pre-trained on the total Places2 training set, ablations among all MAE enhanced methods are still fair and meaningful. For the 512$\times$512 ablations, we finetune the 256$\times$256 model trained on the whole places2 for 150k steps with the dynamic resizing (Sec.~\ref{sec:imp_details}) and test them on 1,000 validation images.

\subsection{Masked Autoencoder for Inpainting}
\label{sec:MAE}

\begin{figure}
\begin{centering}
\includegraphics[width=0.9\linewidth]{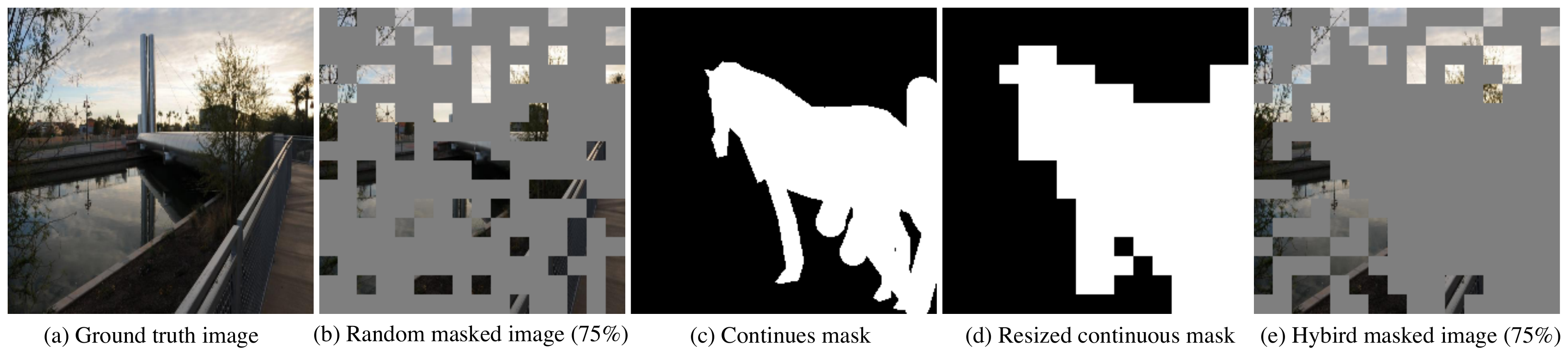}
\par\end{centering}
% \vspace{-0.1in}
\caption{The illustration for masking strategies. In (c), the continuous mask is combined with an irregular mask and a segmentation mask. We upsample the enlarged continues mask (16$\times$16) in (d) for a better visualization.}
% \vspace{-0.15in}
\label{fig:masking_strategy}
\end{figure}

\noindent \textbf{Training Settings}.
We use ViT-Base~\cite{DBLP:journals/corr/abs-2010-11929} as the backbone of MAE, which contains 12 encoder layers and 8 decoder layers. Although He \emph{et al.}~\cite{He2021MaskedAA} have released pre-trained MAEs based on ImageNet-1K with random masks, there are still some domain gaps for the inpainting. Instead, we pre-train the MAE on the whole 1.8 million Places2~\cite{zhou2017places} and 68,000 FFHQ~\cite{8953766} for scene and face inpainting respectively. Validations of both datasets are excepted from the training set for a fair comparison. Moreover, the random mask used in the standard MAE is not amenable to inpainting. Although the masking ratio is high (75\%), such noisy masks are easier to be restored by DNNs compared with continuous and large masks with even lower masking ratios~\cite{ntavelis2020aim}. Therefore, we blend continuous masks with 10\% to 50\% masking ratios and random masks, while the total masking rate remains at 75\% as shown in Fig.~\ref{fig:masking_strategy}. Specifically, both irregular and segmentation masks~\cite{cao2021learning} are considered in continuous masks. Then we downsample continuous masks to 16$\times$16 patch-wise forms with \emph{max pooling} to ensure all masked patches are set to 1, while unmasked ones are 0, which enlarges masked regions. The mixed masking strategy can improve the learning of prior attention as shown in Tab.~\ref{table:MAE_att} of Sec.~\ref{sec:prior_attention}. Other training details are discussed in Sec.~\ref{sec:imp_details}.

\noindent \textbf{Prior Features from MAE}.
In MAE~\cite{He2021MaskedAA}, only 25\% unmasked patches are applied into the encoder, while learnable masked tokens are shared in the decoder for the reconstruction. These masked tokens will be used to predict pixel values in masked regions. This trick makes the encoder enjoy much lower memory cost, and it can be pre-trained with more capacities for better performance in classification tasks. However, during the inpainting, features from the MAE encoder are insufficient. Because we should achieve masked features encoded by masked tokens with stacked attention modules. Therefore, features from decoder layers are more compatible with the inpainting task. To ensure good decoder learning, we chose to use balanced encoder-decoder ViT-Base architecture, which contains 12 encoder layers and 8 decoder layers rather than further enlarging the encoder capacity. In this work, we choose to use features from the last layer of the MAE decoder before the pixel linear projection as the prior features. Since the predicted images are blurry, which may contain limited information, especially for the HR inpainting. Thus an interesting future work would be exploring features from different transformer layers for inpainting\footnote{In our latter ablations, we find that using the feature from 6th layer can achieve superior performance in FAR.}.
% Discussions about the effects of different transformer layers are the future work, which is not included in this paper.
% On the other hand, few discussions about the feature meaning in vision transformers have been exploited before. We think that different decoder layers learn various information. Specifically, early layers in the transformer decoder enjoy more semantic information learned from the encoder, while later layers tend to represent color textures that are close to the final predicted pixels. To further study the influence of different decoder features, we do some related ablations on the subset of Places2 in Tab.~\ref{}. Besides, we also test the result with the image outputs from MAE instead of decoder features.

\begin{figure}
\begin{centering}
\includegraphics[width=0.9\linewidth]{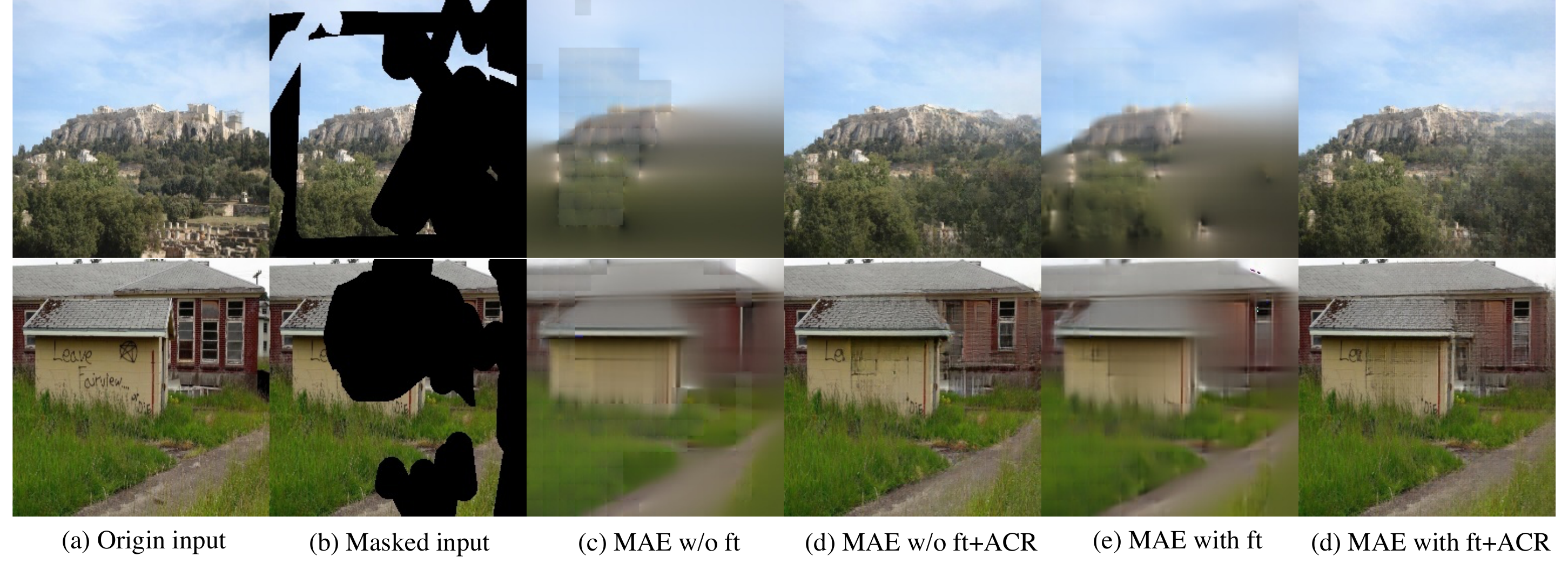}
\par\end{centering}
% \vspace{-0.1in}
\caption{Qualitative results of our full model enhanced by MAE with and without finetuning for partially masked patches.}
% \vspace{-0.15in}
\label{fig:MAE_ft}
\end{figure}

\noindent \textbf{Finetuning for Partially Masked Patches}.
\label{sec:ft_partially}
Since the input of MAE is patch-wise tokens in 16$\times$16, we have to enlarge some partially masked patches as shown in Fig.~\ref{fig:masking_strategy}(c)(d), which may lose information. Intuitively, we further finetune MAE for those partially masked patches with 50 epochs in Places2. Specifically, masked embeddings of these partially masked patches are re-encoded by a new initialized linear layer of decoder. Inputs of these partially masked patches are composed of the concatenation of RGB pixel values (masked pixels are all 0 in 3 channels) and 0-1 masking maps. From the ablations in Tab.~\ref{table:ablation_MAE_ft}, the model trained with finetuned MAE performs slightly worse than the original one in FID. As shown in Fig.~\ref{fig:MAE_ft}, we think that the finetuned MAE learns more explicit results, which makes the CNN restoration overfit MAE features. So these enlarged masked regions increase training difficulty and can be seen as noise regularization.

\begin{table}
% \vspace{-0.1in}
\caption{Ablations of our full model enhanced by MAE with/without finetuning for partially masked patches (Partial F.T.). \label{table:ablation_MAE_ft}}
% \vspace{-0.1in}
\centering
\small
\begin{tabular}{c|cccc}
\hline
Partial F.T. & PSNR$\uparrow$ & SSIM$\uparrow$ & FID$\downarrow$ & LPIPS$\downarrow$\tabularnewline
\hline
$\text{\ensuremath{\times}}$ & 24.51 & \textbf{0.864} & \textbf{25.49} & 0.113 \tabularnewline
$\text{\ensuremath{\checkmark}}$ & \textbf{24.67} & \textbf{0.864} & 26.00 & \textbf{0.111} \tabularnewline
\hline
\end{tabular}
\end{table}

\subsection{Attention-based CNN Restoration (ACR)}
\label{sec:ACR}
The design of CNN modules in ACR is referred to LaMa~\cite{suvorov2021resolution}, which is an encoder-decoder model including 4 downsampling convolutions, 9 Fast Fourier Convolution (FFC) blocks, and 4 upsampling convolutions. FFC has been demonstrated that it can tackle some HR inpainting cases with strong periodic textures. We further enhance ACR with prior features and attentions from MAE as follows.

\noindent \textbf{Prior Features Upsampling}.
\label{sec:upsample}
To overcome inpainting tasks with arbitrary resolutions, the local feature ensemble~\cite{chen2021learning} is leveraged to facilitate the feature warping from MAE to ACR in various resolutions. Given 16$\times$16 patch-wise prior features $\mathbf{F}_p\in\mathbb{R}^{16\times16\times d}$ with dimension $d$ from the MAE decoder, we resize them into 1/8 of the original image size with the bilinear interpolation. To indicate the continuous position in HR, the Cartesian spatial grid~\cite{DBLP:journals/corr/JaderbergSZK15} \emph{i.e.}, normalized 2d coordinates is concated to resized features as
\begin{equation}
\mathbf{F}_p'=\mathrm{Concat}(\mathrm{BilinearResize}(\mathbf{F}_p),\mathbf{C})\in\mathbb{R}^{\frac{h}{8}\times\frac{w}{8}\times (d+2)},
\label{eq:concat_resize_feature}
\end{equation}
where $h,w$ represent the original image height and width respectively; $\mathbf{C}\in\mathbb{R}^{\frac{h}{8}\times\frac{w}{8}\times2}$ means normalized 2d coordinates grid valued from -1 to 1. As shown in the ablations of Tab.~\ref{table:ablation_512}, such positional information can improve the HR inpainting results, and is complementary to learn smooth feature representations. 

\begin{table}
\caption{ \small Ablations of models finetuned in 512$\times$512 Places2 and tested in 1,000 images.  \label{table:ablation_512}}
% \vspace{-0.1in}
\centering
\small
\begin{tabular}{c|c|cccc}
\hline
2D-coordinate & norm-pixel & PSNR$\uparrow$ & SSIM$\uparrow$ & FID$\downarrow$ & LPIPS$\downarrow$\tabularnewline
\hline
$\text{\ensuremath{\times}}$ & $\text{\ensuremath{\checkmark}}$ & 24.35 & 0.879 & 25.47 & 0.135 \tabularnewline
$\text{\ensuremath{\checkmark}}$ & $\text{\ensuremath{\checkmark}}$ & 24.31 & \textbf{0.880} & 25.33 & 0.135 \tabularnewline
$\text{\ensuremath{\checkmark}}$ & $\text{\ensuremath{\times}}$ & \textbf{24.37} & \textbf{0.880} & \textbf{25.30} & \textbf{0.132} \tabularnewline
\hline
\end{tabular}
% \vspace{-0.15in}
\end{table}

\begin{figure}
\begin{centering}
\includegraphics[width=0.7\linewidth]{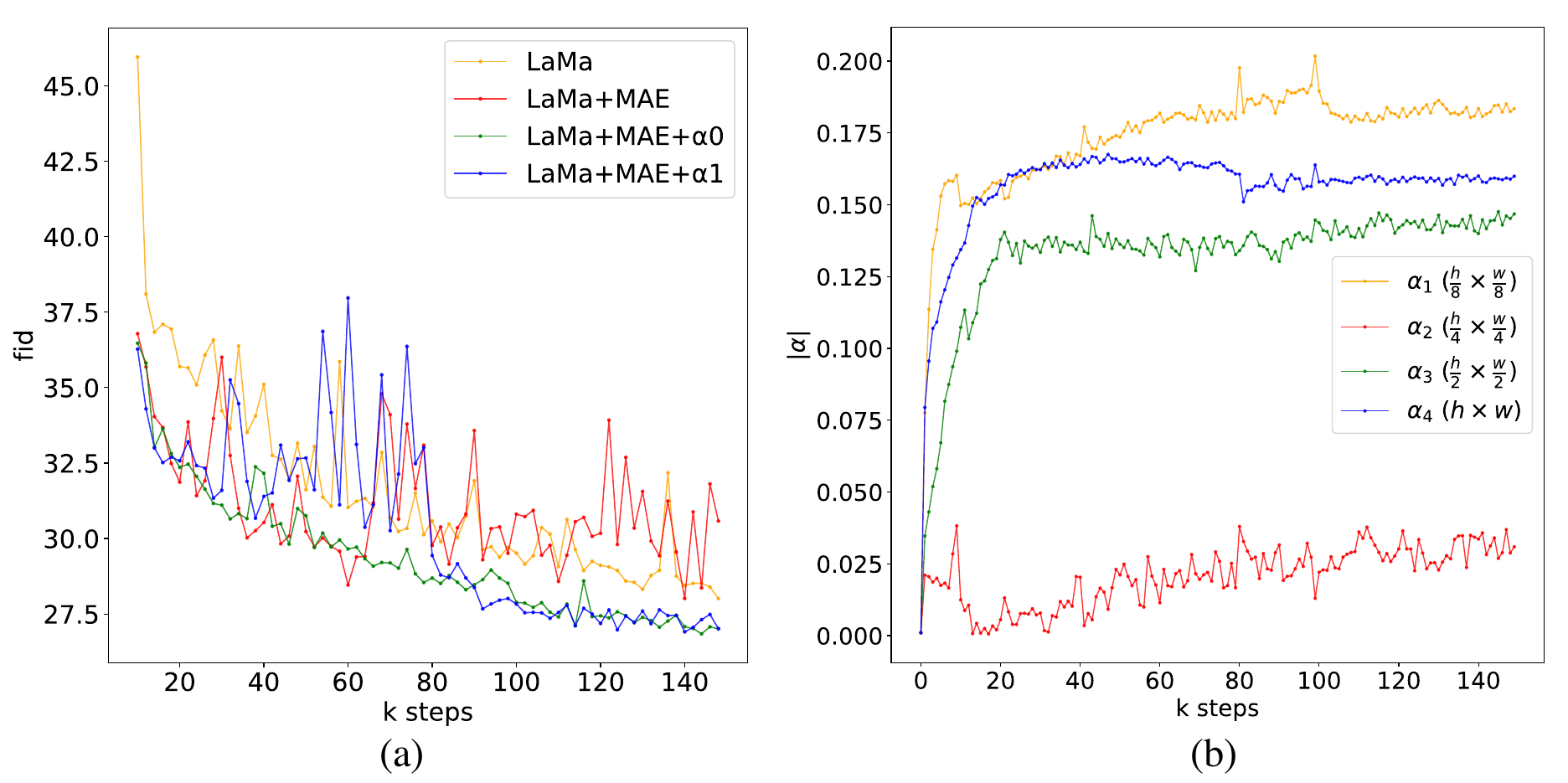}
\par\end{centering}
% \vspace{-0.15in}
\caption{(a) Ablations of various prior features combination methods. LaMa~\cite{suvorov2021resolution} is the baseline. `LaMa+MAE' means directly adding MAE prior features to the ACR encoder. `+$\alpha0$' and `+$\alpha1$' indicate that prior features are multiplied with 0 and 1 initialized learnable-parameter $\alpha$ before the addition. (b) The line chart of $\alpha_j, j\in\{1,2,3,4\}$ shows absolute values for ACR features with different resolutions.}
\label{fig:MAE_alpha_ablation}
% \vspace{-0.2in}
\end{figure}

\noindent \textbf{Prior Features Combination}.
Four gated deconvolutions~\cite{yu2019free} are leveraged to upsample $\mathbf{F}_p'$. Then these upsampled features are applied to the encoder of ACR. The gated mechanism works for making ACR filter corrupted features adaptively. To integrate prior features to ACR, the general solution is to element-wise add upsampled prior features to the downsampled ones of ACR. However, we observed an unstable training process with the vanilla element-wise addition as shown in the ablations of Fig.~\ref{fig:MAE_alpha_ablation}(a). We try to multiply trainable parameters $\alpha_j, j\in\{1,2,3,4\}$ to the prior features for the element-wise addition to ACR encoder features in $(\frac{h}{8}\times\frac{w}{8})$, $(\frac{h}{4}\times\frac{w}{4})$, $(\frac{h}{2}\times\frac{w}{2})$, and $(h\times w)$ respectively. Moreover, ablation studies of $\alpha_j$ initialized with 0 and 1 are shown in Fig.~\ref{fig:MAE_alpha_ablation}(a) and Tab.~\ref{table:ablation_layer_alpha}. The zero initialization enjoys a much more stable convergence. Therefore, the 0-initialized addition is adopted in our model, which is omitted in the follows for simplicity.
We further analyze tendencies of different $\alpha_j$ trained with MAE layer 8 (\emph{i.e.}, row 3 of Tab.~\ref{table:ablation_layer_alpha}) in Fig.~\ref{fig:MAE_alpha_ablation}(b).
From Fig.~\ref{fig:MAE_alpha_ablation}(b), both high-level $(\frac{h}{8}\times\frac{w}{8})$ and low-level $(h\times w,  \frac{h}{2}\times\frac{w}{2})$ features are important for the prior learning, since absolute values of $\alpha_1, \alpha_3, \alpha_4$ are large. And features in $\frac{h}{4}\times\frac{w}{4}$ seem have less effect during the inpainting.

% \noindent \textbf{Prior Features Selection}
% Besides, we share the gated convolutions and train the ACR with prior features from three MAE layers (2, 5, 8) at the same time. Related results are listed in the last row of Tab.~\ref{table:ablation_layer_alpha}. No significant improvement is achieved by using multiple MAE layers, which means that different transformer layers may share similar semantic information.
% corresponds to a feature map size of $\frac{h}{8}\times\frac{w}{8}$ learns faster and better than other $\alpha_j$ across all transformer layers, and $\alpha_2$ is close behind. This is due to they are closer to the output end of the network and play a more important role in final image inpainting results. Besides, we find that there is no improvement when using three transformer layers together, which may be attributed to that different transformer layers share approximately the same semantic information.

\begin{table}
% \vspace{-0.1in}
\caption{Ablations of models enhanced with different initialized parameters $\alpha$. Column `MAE' means that whether to use prior features from MAE. Column `init-$\alpha$' indicates initialized values of learnable parameter $\alpha$ for  prior feature combination.
\label{table:ablation_layer_alpha}}
% \vspace{-0.1in}
\centering
\small
\begin{tabular}{c|c|cccc}
\hline
MAE & init-$\alpha$ & PSNR$\uparrow$ & SSIM$\uparrow$ & FID$\downarrow$ & LPIPS$\downarrow$\tabularnewline
\hline
$\text{\ensuremath{\times}}$ & no & 24.12 & 0.859 & 28.01 & 0.124\tabularnewline
$\text{\ensuremath{\checkmark}}$ & no & 24.14 & 0.860 & 28.01 & 0.127\tabularnewline
$\text{\ensuremath{\checkmark}}$ & 0 & \textbf{24.34} & 0.860 & \textbf{26.83} & 0.117\tabularnewline
$\text{\ensuremath{\checkmark}}$ & 1 & 24.27 & \textbf{0.861} & 26.90 & \textbf{0.115}\tabularnewline
\hline 
\end{tabular}
% \vspace{-0.15in}
\end{table}

% \begin{figure}
% \begin{centering}
% \includegraphics[width=0.85\linewidth]{pics/multi_layer.pdf}
% \par\end{centering}
% \caption{Evolution of absolute $\alpha_j, j\in\{1,2,3,4\}$ across three transformer decoder layers when using them together for ACR.}
% \label{fig:MAE_multi_layer_alpha_ablation}
% \end{figure}

\noindent \textbf{Prior Attentions}.
\label{sec:prior_attention}
Many inpainting researches show that the attention mechanism is useful for the image inpainting~\cite{yu2018generative,yu2019free,zhou2020learning,yi2020contextual}. 
The classical contextual attention~\cite{yu2018generative} used in inpainting is to aggregate masked features $\mathbf{F}_m$ with unmasked ones $\mathbf{F}_u$. The aggregation is based on the attention score $\mathbf{R}_{u,m}$ as
\begin{equation}
\begin{split}
\mathrm{cos}_{u,m}&=\left<\frac{\mathbf{F}_u}{||\mathbf{F}_u||},\frac{\mathbf{F}_m}{||\mathbf{F}_m||}\right>\\
\mathbf{R}_{u,m}&=\mathrm{softmax}_{u}(\mathrm{cos}_{u,m}),
\end{split}
\label{eq:contextual_attention}
\end{equation}
where $cos_{u,m}$ indicates normalized cosine similarities of masked and unmasked features; $\mathrm{softmax}_u$ means the softmax normalization among all unmasked features. Then we can get the aggregated masked features $\mathbf{F}_m'=\sum_{u}\mathbf{R}_{u,m}\mathbf{F}_u$.

Unfortunately, the improvement of the attention module is not orthogonal to other effective inpainting strategies. We add two contextual attention (CA) modules~\cite{yu2018generative} to ACR in the same positions as the prior feature aggregation shown in Fig.~\ref{fig:overview}. But no improvement is achieved by the trainable contextual attention as shown in Tab.~\ref{table:MAE_att}. We think that the restricted improvement is caused by the limited capacity of CNN models. Therefore, we try to use attention relations from the decoder of MAE to overcome the limited capacity. For the decoder layer $l$ of MAE, we can get attention scores $\mathbf{R}_{u,m}^{(l)}$ as
\begin{equation}
\mathbf{R}_{u,m}^{(l)}=\mathrm{softmax}(\frac{\mathbf{Q}^{(l)}\mathbf{K}^{{(l)}^T}}{\sqrt{d}}-inf\cdot \mathbf{M}),
\label{eq:MAE_att}
\end{equation}
where $\mathbf{Q}^{(l)},\mathbf{K}^{(l)}$ mean query and key of the attention; $d$ is the channels; and $\mathbf{M}$ indicates a 0-1 mask map, where 1 means masked regions. We make the scores of masked regions to 0 in Eq.~(\ref{eq:MAE_att}) that means all masked regions should only pay attention to unmasked ones. Then we average values of total 8 decoder attention scores, and get the prior attention scores $\mathbf{R}_p$ as 
\begin{equation}
\mathbf{R}_p=\frac{\sum_{l=1}^{L}\mathbf{R}_{u,m}^{(l)}}{L}, L=8.
\label{eq:prior_att}
\end{equation}
The aggregations for getting masked features $\mathbf{F}_m'$ in ACR are executed in the start and the end of FFC blocks as shown in Fig.~\ref{fig:overview}.
Then $\mathbf{F}_m'$ is added to the original features as the residual. Note that we also multiply a zero-initialized learnable parameter to the $\mathbf{F}_m'$ before the addition instead of using LayerNorm as discussed in~\cite{bachlechner2020rezero}. 
Besides, from Tab.~\ref{table:MAE_att}, random masking used in the vanilla MAE fails to learn proper attention relations compared with mixed one discussed in Sec.~\ref{sec:MAE}. The ablation study about different attention layers from MAE are discussed in the Appendix.

\begin{table}
\small
\caption{Ablations of models with different attention and MAE masking strategies; `mixed' mask type means MAE pre-trained with both random, enlarged irregular and segmentation masks; `prior attention' means using prior attention aggregation from MAE; `trainable CA' indicates contextual attention~\cite{yu2018generative}. 
\label{table:MAE_att}}
\centering
\begin{tabular}{c|c|cccc}
\hline
MAE mask type & attention type & PSNR$\uparrow$ & SSIM$\uparrow$ & FID$\downarrow$ & LPIPS$\downarrow$\tabularnewline
\hline
mixed & no attention & 24.34 & 0.860 & 26.84 & 0.117 \tabularnewline
mixed & trainable CA & 24.13 & 0.859 & 26.99 & 0.123 \tabularnewline
random & prior attention & 24.39 & 0.861 & 26.25 & 0.117 \tabularnewline
mixed & prior attention & \textbf{24.51} & \textbf{0.864} & \textbf{25.49} & \textbf{0.113} \tabularnewline
\hline
\end{tabular}
\end{table}

\subsection{Loss Functions}

\noindent \textbf{Loss Functions of MAE}.
Our MAE loss is the mean squared error (MSE) between MAE predictions and ground truth pixel values for masked patches as in~\cite{He2021MaskedAA}. Besides, He \emph{et al.} study to use normalized pixel values of each masked patch as the self-supervised target, which normalizes each masked patch with the mean and standard of this patch. Such a trick can improve the classification quality in their experiments. However, for the inpainting, the global relation among different patches is also important. The patch-wise normalization makes MAE learn more bias for each patch rather than the global information.
Thus we study the MAE pre-training ablations with and without the patch-wise normalization as shown in Tab.~\ref{table:ablation_512}. We find that non-normalized targets can achieve slightly better quality in HR inpainting.

\noindent \textbf{Loss Functions of ACR}.
ACR is trained as a regular adversarial inpainting model. We adopt the same loss functions as LaMa~\cite{suvorov2021resolution}, which include L1 loss, adversarial loss, feature match loss, and high receptive field (HRF) perceptual loss. Specifically, L1 loss is only used to constrain unmasked regions. The discriminator loss is based on PatchGAN~\cite{isola2017image}. And WGAN-GP~\cite{gulrajani2017improved} is used as the generator loss. The feature match loss~\cite{wang2018high} based on L1 loss between true and fake discriminator features is also used to stable the GAN training. Furthermore, we also leverage the segmentation pre-trained ResNet50 for the HRF loss as proposed in~\cite{suvorov2021resolution} to improve the inpainting quality. More details about the loss functions of ACR are in the Appendix.

\section{Experiments}

\noindent \textbf{Datasets}.
Our model is trained on Places2~\cite{zhou2017places} and FFHQ~\cite{8953766}. For Places2, both the pre-training of MAE and training of our full model are based on the whole 1.8 million training images, and tested with 36,500 testing images. We prepare additional 1,000 testing images for 512$\times$512 experiment.
Detailed settings about main experiments and ablations are in the Appendix.
We pre-train another MAE on 68,000 training set of FFHQ, which is also used to train the face inpainting model. And other 2,000 images work for testing.
Our pre-trained MAE on Places2 and FFHQ can be well generalized to most real-world cases.

%about main experiments and ablations of Places2 are reorganized 
%which are sufficient for the inpainting research.
%In addition, to make ablation experiments easier, we created a tiny dataset (P2T\footnote{Details about P2T are illustrated in the appendix.}) out of 25,500 indoor and outdoor photos from the Plces2 dataset, with 500 of them serving as the test set. 
%We train on both Places2 and FFHQ for input images that support $256\times256$ and $512\times512$ resolutions.

\begin{table}
\centering
\caption{\textbf{Left:} Results on 256$\times$256 FFHQ and Places2 with mixed masks compared among Co-Mod (Co.)~\cite{zhao2021large}, LaMa (La.)~\cite{suvorov2021resolution}, EC~\cite{nazeri2019edgeconnect}, CTSDG (CT.)~\cite{Guo_2021_ICCV} and ours. 
\textbf{Right:} Results on 512$\times$512 Places2 testset with mixed masks compared among HiFill (Hi.)~\cite{yi2020contextual}, Co-Mod (Co.)~\cite{zhao2021large}, LaMa (La.)~\cite{suvorov2021resolution} and ours.
Metrics are PSNR (P.), SSIM (S.), FID (F.) and LPIPS (L.).}
\setlength{\tabcolsep}{1mm}{
\begin{tabular}{cc}
\small
\label{table:main_results}
\begin{tabular}{c|ccc|ccccc}
\hline
 & \multicolumn{3}{c|}{FFHQ} & \multicolumn{5}{c}{Places2}\tabularnewline
\hline
 & Co. & La. & Ours & EC & Co. & La. & CT. & Ours\tabularnewline
\hline
\multirow{1}{*}{P.$\uparrow$} & 25.2 & 26.6 & \textbf{26.8} & 23.3 & 22.5 & 24.3 & 23.4 & \textbf{24.5}\tabularnewline
\multirow{1}{*}{S.$\text{\ensuremath{\uparrow}}$} & .889 & .903 & \textbf{.906} & .839 & .843 & .869 & .835 & \textbf{.871}\tabularnewline
\multirow{1}{*}{F.$\downarrow$} & \textbf{5.85} & 6.38 & 6.15 & 6.21 & 1.49 & 1.63 & 11.2 & \textbf{1.31}\tabularnewline
\multirow{1}{*}{L.$\downarrow$} & .085 & .078 & \textbf{.074} & .149 & .246 & .155 & .185 & \textbf{.101}\tabularnewline
\hline
\end{tabular}
& 
\begin{tabular}{c|cccc}
\hline
 & P.$\uparrow$ & S.$\text{\ensuremath{\uparrow}}$ & F.$\downarrow$ & L.$\downarrow$\tabularnewline
\hline
Hi. & 20.1 & .764 & 65.4 & .291\tabularnewline
Co. & 22.0 & .843 & 30.0 & .166\tabularnewline
La. & 24.1 & .877 & 27.8 & .149\tabularnewline
Ours & \textbf{24.3} & \textbf{.880} & \textbf{25.3} & \textbf{.119}\tabularnewline
\hline
\end{tabular}\tabularnewline
\end{tabular}}
\end{table}

\begin{figure}
\begin{centering}
\includegraphics[width=0.9\linewidth]{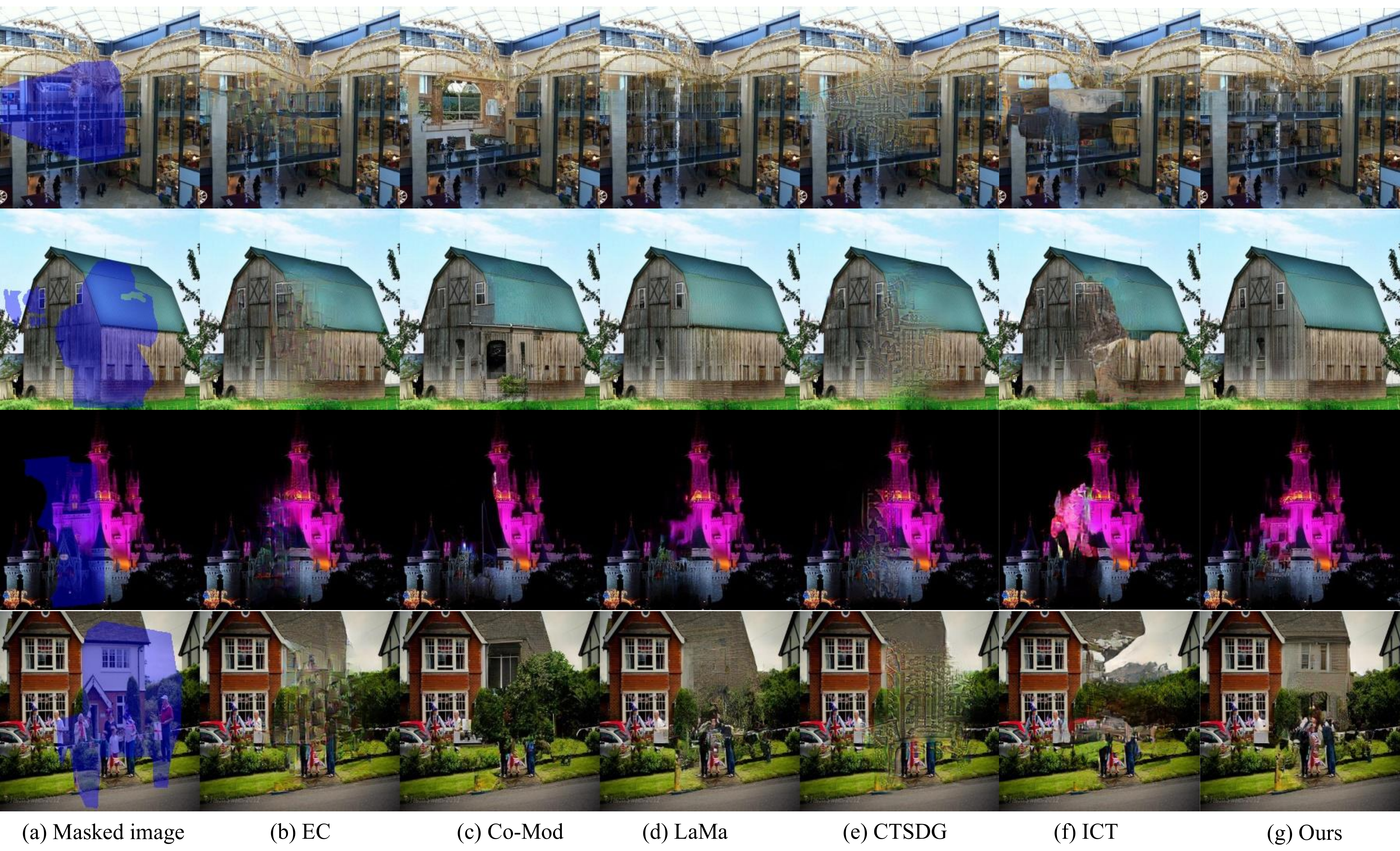}
\par\end{centering}
% \vspace{-0.15in}
\caption{Qualitative results of places2 256$\times$256 images. From left to right are masked image, EC~\cite{nazeri2019edgeconnect}, Co-Mod~\cite{zhao2021large}, LaMa~\cite{suvorov2021resolution}, CTSDG~\cite{Guo_2021_ICCV}, ICT~\cite{wan2021highfidelity}, and our results.}
% \vspace{-0.15in}
\label{fig:qualitative_places2_256}
\end{figure}

\begin{figure}
 \centering 
 \includegraphics[width=0.99\linewidth]{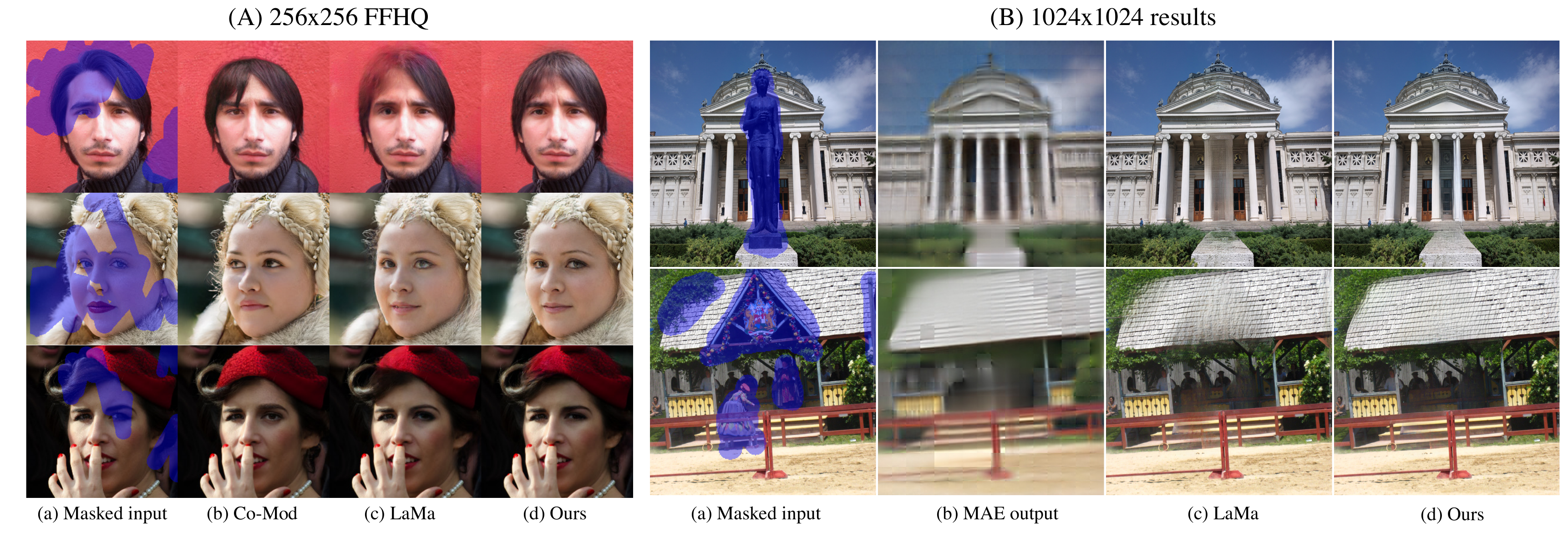}
 \caption{Qualitative (A) 256$\times$256 FFHQ, and (B) 1024$\times$1024 results from network.}
 \label{fig:face_and_highres}
\end{figure}

\noindent \textbf{Implementation Detials}.
\label{sec:imp_details}
Our FAR is implemented with Pytorch based on two 48 GB NVIDIA RTX A6000 gpus. MAE is pre-trained on Places2 and FFHQ for 200 and 450 epochs respectively with batch size 512, while other settings follow the released codes\footnote{\url{https://github.com/facebookresearch/mae}} except the masking strategy and the partially masked finetuning. For ACR, we employ the Adam optimizer with $\beta_1=0.9$ and $\beta_2=0.999$, the learning rates are 1e-3 and 1e-4 for the generator and the discriminator. We train our models with 850k steps on Places2 and 150k steps on FFHQ. For every 200k steps on Places2 and 100k steps on FFHQ, the learning rate is reduced by half. 
To save the computation, we finetune our model on images with higher resolutions, which are dynamically resized from 256 to 512 for 150k steps on Places2. 
% For the FFHQ, we finetune our model with 512 for 100k steps.

\noindent \textbf{Masks Settings}.
To solve real-world application problems, we adopt the masking strategy in~\cite{cao2021learning}. The masks consist of irregular brush masks and COCO~\cite{lin2014microsoft} segmentation masks ranged from 10\% to 50\%. During the training, we combine these two types masks in 20\%.

\noindent \textbf{Comparison Methods}.
We compared our model with other state-of-the-art models including Edge Connect (EC)~\cite{nazeri2019edgeconnect}, Co-Modulation GAN
(Co-Mod)~\cite{zhao2021large}, Large Mask inpainting (LaMa)~\cite{suvorov2021resolution}, Conditional Texture and Structure Dual Generation (CTSDG)~\cite{Guo_2021_ICCV} and Image Completion with Transformers (ICT)~\cite{wan2021highfidelity} using the official pre-trained models. We also retrain LaMa on Places2 and FFHQ, and further finetune it with the same settings as ours in the HR inpainting for a fair comparison.

\begin{figure}
\begin{centering}
\includegraphics[width=0.99\linewidth]{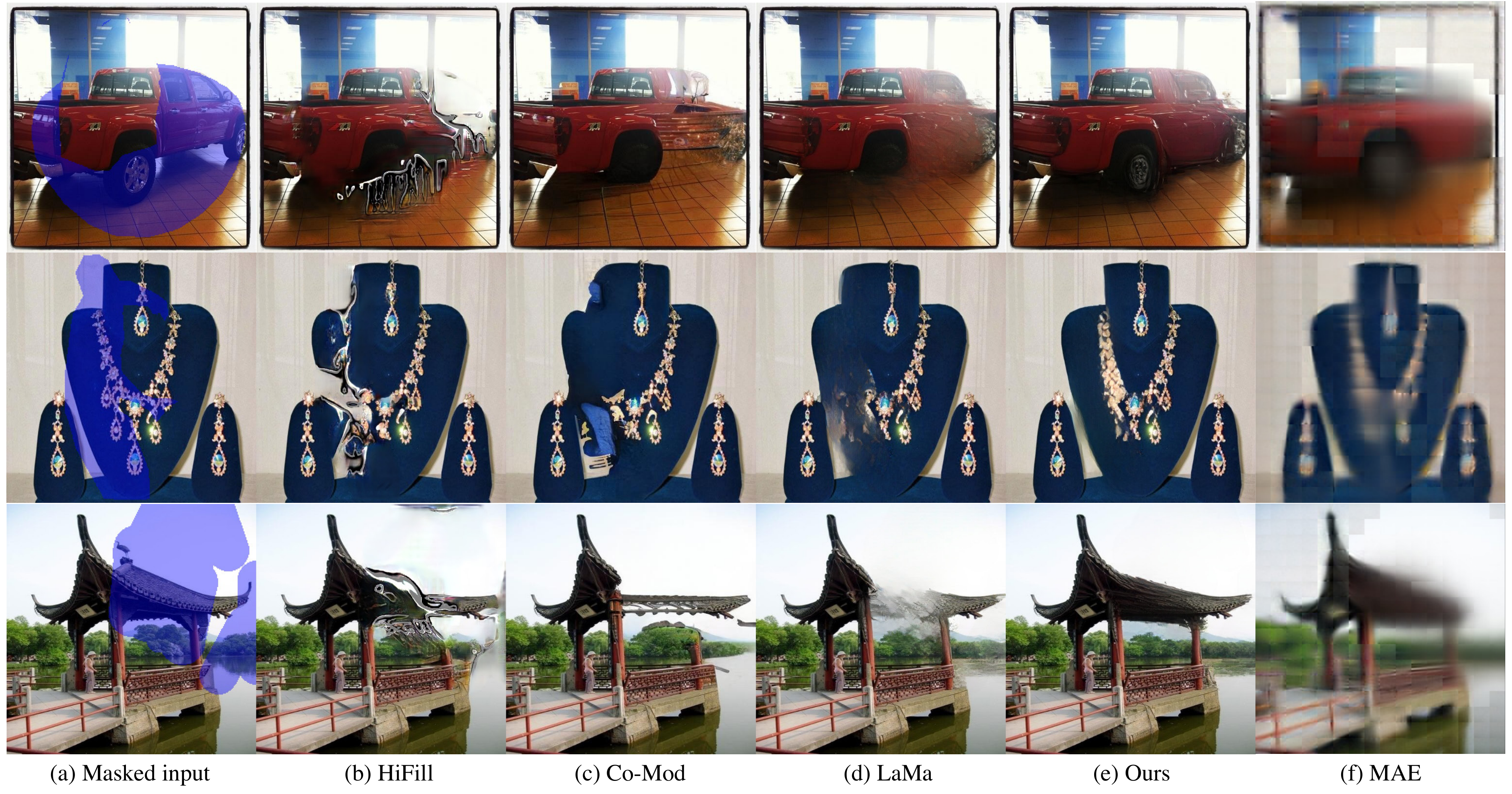}
\par\end{centering}
\caption{Qualitative results of places2 512$\times$512 images. From left to right are masked image, HiFill~\cite{yi2020contextual}, Co-Mod~\cite{zhao2021large}, LaMa~\cite{suvorov2021resolution}, our results, and MAE predictions.\label{fig:qualitative_places2_512}}
\end{figure}

% \subsection{Image Inpainting Results}
\noindent \textbf{Quantitative Comparisons}.
We evaluate PSNR, SSIM~\cite{wang2004image}, FID~\cite{heusel2018gans}, and LPIPS~\cite{zhang2018unreasonable} for both 256$\times$256 and 512$\times$512 in Tab.~\ref{table:main_results}. 
For 256$\times$256 results shown in the left of Tab.~\ref{table:main_results}, our method achieves significant improvements based on the LaMa baseline. Moreover, Co-Mod results are also competitive. 
For the FFHQ results, since our MAE enhanced method can achieve much more stable and faithful results for face images, PSNR and SSIM of our method is better than Co-Mod. The powerful stylegan~\cite{karras2020analyzing} architecture helps Co-Mod learn better textures with good FID, but our results have superior human perception (lower LPIPS). For the Places2, our results achieve best results in all metrics.

For the HR 512$\times$512 results listed in the right of Tab.~\ref{table:main_results}, our method can outperform all other competitors. Note that most methods fail to tackle HR inpainting tasks in complex scenes. Results of baseline method LaMa are competitive, but our methods still achieve certain advantages compared with LaMa.

\begin{figure}
\begin{centering}
\includegraphics[width=0.99\linewidth]{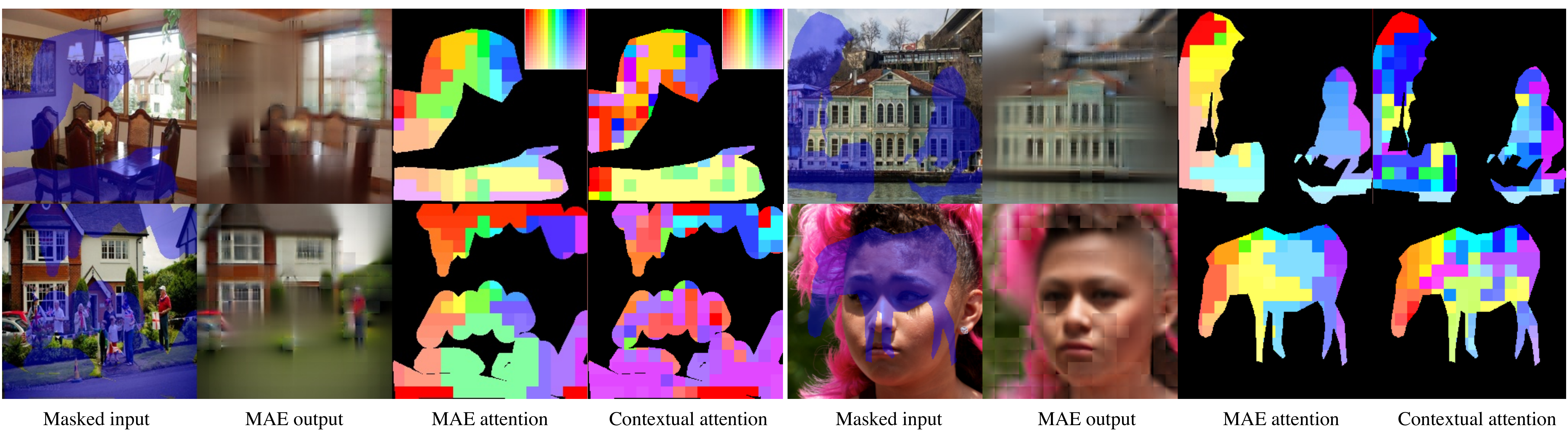}
\par\end{centering}
\caption{Visualization of attentions from MAE and learned by  contextual attention mechanism~\cite{yu2018generative}. For the visualizations of attention maps, unmasked regions are ignored as black, and the patch attended with the highest attention score is shown in masked regions. Attention palettes are shown in the upper right corner of the first instance, which illustrate the exact positions attended mostly by masked patches.}
\label{fig:vis_att}
\end{figure}

\begin{table}
\caption{Efficiency of our model without  pre-processing.  Results are reported on 256$\times$256 images with training and testing batch size 24 and 1 respectively. \label{table:time}}
% Both training and inference are  256$\times$256. 
% Training is tested with batch size 24, while the inference is tested with batch size 1. }
% \vspace{-0.1in}
\centering
\small
\begin{tabular}{c|c|c|c}
\hline 
MAE  & Attention & Training(sec/batch) & Inference(sec/image)\tabularnewline
\hline 
$\text{\ensuremath{\times}}$ & $\text{\ensuremath{\times}}$ & 1.0101 & 0.0425 \tabularnewline
$\text{\ensuremath{\checkmark}}$ & $\text{\ensuremath{\times}}$ & 1.2510 & 0.0656 \tabularnewline
$\text{\ensuremath{\checkmark}}$ & $\text{\ensuremath{\checkmark}}$ & 1.2604 & 0.0665
\tabularnewline
\hline 
\end{tabular}
% \vspace{-0.25in}
\end{table}

\noindent \textbf{Qualitative Comparisons}. 
We show 256$\times$256 qualitative results of Places2 and FFHQ in Fig.~\ref{fig:qualitative_places2_256} and Fig.~\ref{fig:face_and_highres}(A) respectively. For the results in Fig.~\ref{fig:qualitative_places2_256}, results of EC and CTSDG are blurry and have obvious artifacts. The results of Co-Mod and LaMa have proper qualities. But these two methods generate some unreasonable building architectures, which also cause artifacts. For the results of ICT, they fail to get good quality due to the poor LR reconstruction. Moreover, the slow sampling strategy of ICT makes it hard to test large-scale image datasets for a quantitative comparison. Our method can achieve better inpainting results in both structures and textures. For face images, our method can achieve stable inpainted results with consistent eye gaze.

For qualitative results in 512$\times$512, other compared algorithms fail to reconstruct reasonable semantics for certain objects in HR, such as car, necklace, and temple in Fig.~\ref{fig:qualitative_places2_512}, except our FAR. Such good results are benefited from informative prior features of MAE, which make the model understand the real categories of objects in HR cases. Besides, as shown in Fig.~\ref{fig:qualitative_places2_512}, our method avoids overfitting MAE outputs due to the enlarged masking strategy discussed in Sec.~\ref{sec:ft_partially}. We further provide some 1024 results in Fig.~\ref{fig:face_and_highres}(B), which show that MAE can provide expressive priors for both structure and texture reconstructions.

% \subsection{Visualization of Attention}
% \label{sec:vis_att}

\noindent \textbf{Visualization of Attention}.
\label{sec:vis_att}
We show the visualization about the attention scores in Fig.~\ref{fig:vis_att}. Patches attended with the highest attention score are shown in MAE and contextual attention maps, which can be located by the palettes in the first instance of Fig.~\ref{fig:vis_att}. Attention scores from MAE are more consistent and reasonable compared with learning-based contextual attention~\cite{yu2018generative}. Specifically, some irrelevant patches are attended by masked regions in the contextual attention, which leads to confused attention maps. Besides, although some MAE results are blurry, their attention relations are still effective.

\noindent \textbf{Computation and Complexity}. Benefited by the efficient design from~\cite{He2021MaskedAA} with 25\% unmasked input tokens to the encoder, we should claim that the MAE pre-training is not heavier than CNNs. As discussed in implemented details, our MAE has been trained for 200 epochs in Places2 with about 10 days on two NVIDIA RTX A6000 gpus, which costs almost the same time for training a LaMa in Places2 for just 800k steps (only 28.4 epochs). Besides, we list times for both training and inference stages tested on A6000 in Tab.~\ref{table:time}. It shows that the computation of the prior attention is negligible and can be ignored. MAE and gated convolutions increase about 0.024 seconds for predicting each image, which affects the training a little compared with the time-consuming GAN training. And the inference with 0.0665 seconds for each image is efficient enough for the user interaction, \emph{e.g.}, object removal.

\section{Conclusions}

This paper proposes an MAE enhanced image inpainting model called FAR. We utilize a masked visual prediction based vision transformer -- MAE to provide features for the CNN based inpainting model, which contain rich informative priors with meaningful semantics. Moreover, we apply the prior attention scores from the pre-trained MAE to aggregate masked features, which is proved to work better than learning contextual attention from scratch. Besides, many constructive issues about the pre-trained MAE and image inpainting are discussed in this paper. Our experiments show that our method can achieve good improvements with the prior features and attentions from MAE. Social impacts of our model, especially working on the face dataset are discussed in the Appendix.\\

\noindent\textbf{Acknowledgements}. This work was supported by SMSTC (2018SHZDZX01).

\appendix

\section{Social Impact}

All generated results of both the main paper and the appendix are based on learned statistics of the training dataset. Therefore, the results only reflect biases in those released data without our subjective opinion, especially for the face images from FFHQ. This work is only researched for the algorithmic discussion, and related societal impacts should not be ignored by users. 

\section{Detailed Network Settings}
We provide some details for different model components in this section.
\noindent\textbf{Gated Convolution Block (GC)}. For the GC block used for upsampling prior features from MAE, which contains GateConv2D~\cite{yu2019free}$\rightarrow$BatchNorm$\rightarrow$ReLU. And the GateConv2D works with stride=2.

\noindent\textbf{Encoder and Decoder of ACR}. The encoder and decoder of ACR are consisted of vanilla Conv2D$\rightarrow$BatchNorm$\rightarrow$ReLU.

\noindent\textbf{Fast Fourier Convolution Block (FFC)}. As illustrated in~\cite{suvorov2021resolution}, features for FFC are split into local ones encoded by vanilla convolutions and global ones encoded by the spectral transform. The spectral transform is consisted of Fast Fourier Transform (FFT), Conv2D$\rightarrow$BatchNorm$\rightarrow$ReLU, and the inverse FFT. And both the real and imaginary parts are confirmed in the Conv2D after FFT. After the inverse FFT, local and global features are combined as the final output.

\section{Loss Functions of ACR}
We provide some details about the loss functions of ACR, which are referred to LaMa~\cite{suvorov2021resolution} and include L1 loss, adversarial loss, feature match loss, and high receptive field (HRF) perceptual loss~\cite{suvorov2021resolution}.
L1 loss is only calculated between the unmasked regions as
\begin{equation}
\mathcal{L}_{L1}=(1-\mathbf{M})\odot|\mathbf{\hat{I}}-\mathbf{\tilde{I}}|_1,
\label{eq:l1_loss}
\end{equation}
where $\mathbf{M}$ indicates 0-1 mask that 1 means masked regions; $\odot$ means the element-wise multiplication; $\mathbf{\hat{I}},\mathbf{\tilde{I}}$ indicate the ground truth and predicted images respectively. The adversarial loss is consisted of a PatchGAN~\cite{isola2017image} based discriminator loss $\mathcal{L}_D$ and a WGAN-GP~\cite{gulrajani2017improved} based generator loss $\mathcal{L}_G$ as
\begin{equation}
\begin{split}
\mathcal{L}_{D}=&-\mathbb{E}_{\mathbf{\hat{I}}}\left[\mathrm{log}D(\mathbf{\hat{I}})\right]-\mathbb{E}_{\mathbf{\tilde{I}},\mathbf{M}}\left[\mathrm{log}D(\mathbf{\tilde{I}})\odot(1-\mathbf{M})\right]\\
&-\mathbb{E}_{\mathbf{\tilde{I}},\mathbf{M}}\left[\mathrm{log}(1-D(\mathbf{\tilde{I}}))\odot\mathbf{M}\right],\\
&\mathcal{L}_{G}=-\mathbb{E}_{\mathbf{\tilde{I}}}\left[\mathrm{log}D(\mathbf{\tilde{I}})\right],\\
&\mathcal{L}_{adv}=\mathcal{L}_D+\mathcal{L}_G+\lambda_{GP}\mathcal{L}_{GP},
\end{split}
\label{eq:adv_loss}
\end{equation}
where $\mathcal{L}_{GP}=\mathbb{E}_{\mathbf{\hat{I}}}||\triangledown_{\mathbf{\hat{I}}}D(\mathbf{\hat{I}})||^2$ is the gradient penalty~\cite{gulrajani2017improved} and $\lambda_{GP}=1e-3$. Moreover, the feature match loss~\cite{wang2018high} $\mathcal{L}_{fm}$, which is based on L1 loss between discriminator features $D_f$ of true and fake samples as 
\begin{equation}
\mathcal{L}_{fm}=\mathbb{E}(|D_f(\hat{\mathbf{I}})-D_f(\tilde{\mathbf{I}})|_1).
\label{eq:fm_loss}
\end{equation}
Furthermore, we use the HRF loss $\mathcal{L}_{hrf}$ in~\cite{suvorov2021resolution} as
\begin{equation}
\mathcal{L}_{hrf}=\mathbb{E}(\left[\phi_{hrf}(\mathbf{\hat{I}})-\phi_{hrf}(\mathbf{\tilde{I}})\right]^2),
\label{eq:hrfpl}
\end{equation}
where $\phi_{hrf}$ indicates a pretrained segmentation ResNet50 with dilated convolutions, which shows superior performance in inpainting compared with vanilla VGG as discussed in~\cite{suvorov2021resolution}.
The final loss of our model can be written as
\begin{equation}
\mathcal{L}_{final}=\lambda_{L1}\mathcal{L}_{L1}+\lambda_{adv}\mathcal{L}_{adv}+\lambda_{fm}\mathcal{L}_{fm}+\lambda_{hrf}\mathcal{L}_{hrf},
\label{eq:final_loss}
\end{equation}
where $\lambda_{L1}=10,\lambda_{adv}=10,\lambda_{fm}=100,\lambda_{hrf}=30$ set by the experience.

\section{More Implement Details}
\noindent\textbf{High-Resolution (HR) Finetuning}. To save the computation, we find that dynamic finetune the inpainting model from 256 to 512 resolutions can still achieve competitive results. We gradually reduce the resolution from 512 to 256, and then let them back to 512, which can be seen as a cycle. For each epoch in Places2, we finetune the model with 64 cycles.

\noindent\textbf{The Subset of Places2 for Ablations}. To flexibly evaluate our ablation studies, we choose to use a subset of Places2 with 5 scenes of `bow\_window', `house', `village', `dining\_room', and `viaduct' with about 25,000 training images and 500 validation images. This subset contains indoor, outdoor, and natural scenes, which are comprehensive to evaluate the inpainting performance.

\noindent\textbf{Detailed Settings of Places2}. We reorganize detailed settings of main experiments and ablations in Tab.~\ref{tab:exp_setting}.

\begin{table}[h]
\small
 \centering
 \caption{Settings of main experiments (Main Exp.) and ablations. `Res.', `F.T.' mean resolution and finetuned. Models without finetuning are trained from scratch with pre-trained MAE. `HR subset' indicate validated images of 512$\times$512. Related data scales are in brackets.\label{tab:exp_setting}}
 {\footnotesize{}}%
 \setlength{\tabcolsep}{0.6mm}{
 {\footnotesize{}}%
 \begin{tabular}{c|c|c|c|c|c|c|c}
 \hline 
 \multirow{3}{*}{{\footnotesize{}Exp.}} & \multirow{3}{*}{{\footnotesize{}Res.}} & \multirow{3}{*}{{\footnotesize{}F.T. from}} & \multicolumn{2}{c|}{{\footnotesize{}Places2(train)}} & \multicolumn{3}{c}{{\footnotesize{}Places2(eval)}}\tabularnewline
 \cline{4-8} \cline{5-8} \cline{6-8} \cline{7-8} \cline{8-8} 
  &  &  & {\footnotesize{}whole} & {\footnotesize{}subset} & {\footnotesize{}whole} & {\footnotesize{}subset} & {\footnotesize{}HR subset}\tabularnewline
  &  &  & {\scriptsize{}(1.8M)} & {\scriptsize{}(25,000)} & {\scriptsize{}(36,500)} & {\scriptsize{}(500)} & {\scriptsize{}(1,000)}\tabularnewline
 \hline 
 {\footnotesize{}MAE} & {\footnotesize{}256} & {\footnotesize{}--} & {\footnotesize{}$\checkmark$} &  & {\footnotesize{}--} & {\footnotesize{}--} & {\footnotesize{}--}\tabularnewline
 \hline 
 {\footnotesize{}Main Exp.-1} & {\footnotesize{}256} & {\footnotesize{}--} & {\footnotesize{}$\checkmark$} &  & {\footnotesize{}$\checkmark$} &  & \tabularnewline
 \hline 
 {\footnotesize{}Main Exp.-2} & {\footnotesize{}512} & {\footnotesize{}Main Exp.-1} & {\footnotesize{}$\checkmark$} &  &  &  & {\footnotesize{}$\checkmark$}\tabularnewline
 \hline 
 {\footnotesize{}Ablations-1} & {\footnotesize{}256} & {\footnotesize{}--} &  & {\footnotesize{}$\checkmark$} &  & {\footnotesize{}$\checkmark$} & \tabularnewline
 \hline 
 {\footnotesize{}Ablations-2} & {\footnotesize{}512} & {\footnotesize{}Main Exp.-1} & {\footnotesize{}$\checkmark$} &  &  &  & {\footnotesize{}$\checkmark$}\tabularnewline
 \hline 
 \end{tabular}{\footnotesize\par}}
 \end{table}

\section{User Study}

\begin{figure}
\begin{centering}
\includegraphics[width=0.85\linewidth]{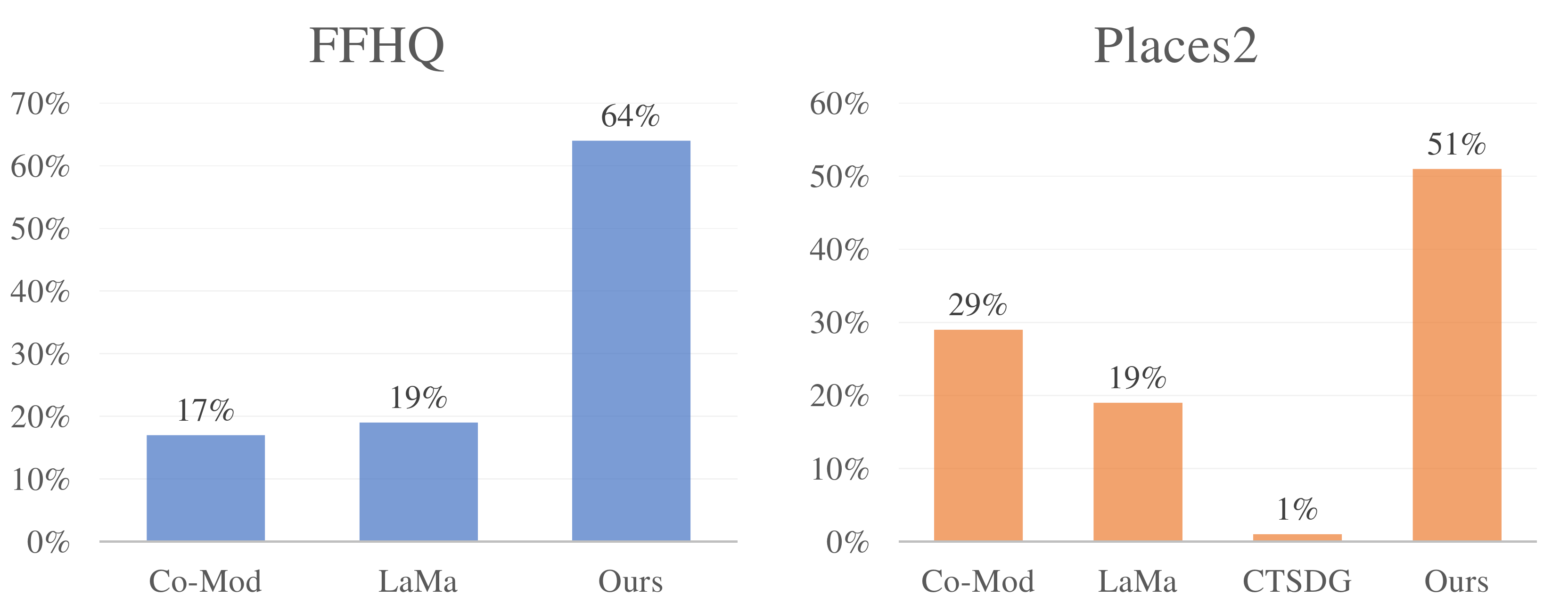}
\par\end{centering}
\caption{Average scores of FFHQ and Places2 for user studies, which are
collected from volunteers who select the best one from shuffled inpainted images.}
\label{fig:user-study}
\end{figure}

To test the effectiveness of our model with human perception, we conduct user studies on several models. We specifically ask 12 participants who are unfamiliar with image inpainting to judge the quality of inpainted images. FFHQ is compared with three models: Co-Mod~\cite{zhao2021large}, LaMa~\cite{suvorov2021resolution} and ours, while Places2 is compared with four methods: Co-Mod, LaMa, CTSDG~\cite{Guo_2021_ICCV} and ours. We randomly shuffle and combine the outcomes of these algorithms except the masked inputs. After that, volunteers must choose the best one from each group. On both datasets, as shown in Fig.~\ref{fig:user-study}, our technique outperforms other competitors. Although Co-Mod also achieves competitive results in Places2, it is trained with extra 6.2 million images from Places365-Challenge, which is much larger than the training set of other competitors.

\section{Complete Quantitative Results}

Tab.~\ref{table:main_results_supp} presents more quantitative results for different masking rates ranging from 10\% to 50\% on datasets FFHQ and Places2. Except for the FID on the FFHQ, our model beats other state-of-the-art methods on other metrics, which demonstrates the superiority of our model.

\begin{table*}
\caption{Quantitative results on FFHQ and Places2 with different mask ratios. \label{table:main_results_supp}}
\centering
\small
\begin{tabular}{c|c|ccc|ccccc}
\hline
 &  & \multicolumn{3}{c|}{FFHQ (256$\times$256)} & \multicolumn{5}{c}{Places2 (256$\times$256)}\tabularnewline
\hline
\hline
 & Mask & Co-Mod & LaMa & Ours & EC & Co-Mod & LaMa & CTSDG & Ours\tabularnewline
\hline
\multirow{5}{*}{PSNR$\uparrow$} & 10\textasciitilde 20\% & 28.45 & 29.84 & \textbf{30.11} & 26.61 & 26.40 & 28.23 & 26.73 & \textbf{28.36}\tabularnewline
 & 20\textasciitilde 30\% & 26.04 & 27.52 & \textbf{27.78} & 24.26 & 23.61 & 25.31 & 24.37 & \textbf{25.48}\tabularnewline
 & 30\textasciitilde 40\% & 24.29 & 25.82 & \textbf{26.07} & 22.60 & 21.73 & 23.44 & 22.71 & \textbf{23.60}\tabularnewline
 & 40\textasciitilde 50\% & 22.93 & 24.48 & \textbf{24.71} & 21.28 & 20.28 & 22.03 & 21.41 & \textbf{22.18}\tabularnewline
 & Mixed & 25.25 & 26.60 & \textbf{26.81} & 23.31 & 22.57 & 24.37 & 23.43 & \textbf{24.53}\tabularnewline
\hline
\multirow{5}{*}{SSIM$\text{\ensuremath{\uparrow}}$} & 10\textasciitilde 20\% & 0.938 & 0.950 & \textbf{0.951} & 0.913 & 0.926 & \textbf{0.942} & 0.913 & \textbf{0.942}\tabularnewline
 & 20\textasciitilde 30\% & 0.909 & 0.924 & \textbf{0.926} & 0.872 & 0.880 & 0.901 & 0.872 & \textbf{0.903}\tabularnewline
 & 30\textasciitilde 40\% & 0.876 & 0.897 & \textbf{0.899} & 0.828 & 0.831 & 0.859 & 0.828 & \textbf{0.861}\tabularnewline
 & 40\textasciitilde 50\% & 0.843 & 0.869 & \textbf{0.872} & 0.783 & 0.781 & 0.814 & 0.782 & \textbf{0.818}\tabularnewline
 & Mixed & 0.889 & 0.903 & \textbf{0.906} & 0.839 & 0.843 & 0.869 & 0.835 & \textbf{0.871}\tabularnewline
\hline
\multirow{5}{*}{FID$\downarrow$} & 10\textasciitilde 20\% & \textbf{3.22} & 3.60 & 3.42 & 1.95 & 0.52 & 0.45 & 2.44 & \textbf{0.41}\tabularnewline
 & 20\textasciitilde 30\% & \textbf{4.66} & 5.20 & 4.94 & 3.79 & 1.00 & 0.95 & 5.62 & \textbf{0.81}\tabularnewline
 & 30\textasciitilde 40\% & \textbf{5.68} & 6.57 & 6.14 & 6.98 & 1.65 & 1.73 & 11.43 & \textbf{1.40}\tabularnewline
 & 40\textasciitilde 50\% & \textbf{7.04} & 8.69 & 8.12 & 11.50 & 2.38 & 2.82 & 19.88 & \textbf{2.20}\tabularnewline
 & Mixed & \textbf{5.85} & 6.38 & 6.15 & 6.21 & 1.49 & 1.63 & 11.18 & \textbf{1.31}\tabularnewline
\hline
\multirow{5}{*}{LPIPS$\downarrow$} & 10\textasciitilde 20\% & 0.049 & 0.045 & \textbf{0.043} & 0.073 & 0.053 & 0.047 & 0.085 & \textbf{0.042}\tabularnewline
 & 20\textasciitilde 30\% & 0.069 & 0.062 & \textbf{0.059} & 0.111 & 0.098 & 0.083 & 0.133 & \textbf{0.073}\tabularnewline
 & 30\textasciitilde 40\% & 0.091 & 0.082 & \textbf{0.077} & 0.152 & 0.140 & 0.121 & 0.185 & \textbf{0.106}\tabularnewline
 & 40\textasciitilde 50\% & 0.113 & 0.101 & \textbf{0.095} & 0.194 & 0.184 & 0.161 & 0.237 & \textbf{0.141}\tabularnewline
 & Mixed & 0.085 & 0.078 & \textbf{0.074} & 0.149 & 0.246 & 0.155 & 0.185 & \textbf{0.101}\tabularnewline
\hline
\end{tabular}
\end{table*}

\section{Quantitative Comparisons on DIV2K}
We further give quantitative high-resolution results on 100 DIV2K~\cite{Agustsson_2017_CVPR_Workshops} validation images with 2k resolutions in Tab.~\ref{tab:div2k}. Following the evaluation protocol of DIV2K in Tab.~\ref{tab:div2k}. Our model beats other HR inpainting methods.

\begin{table}[h]
\small
\centering
\caption{Quantitative results on DIV2K with mixed masks.\label{tab:div2k}}
\begin{tabular}{c|cccc}
\hline
 & {\footnotesize{}PSNR$\uparrow$} & {\footnotesize{}SSIM$\uparrow$} & {\footnotesize{}FID$\downarrow$} & {\footnotesize{}LPIPS$\downarrow$}\tabularnewline
\hline
{\footnotesize{}HiFill} & {\footnotesize{}20.67} & {\footnotesize{}0.787} & {\footnotesize{}135.53} & {\footnotesize{}0.241}\tabularnewline
{\footnotesize{}LaMa} & {\footnotesize{}21.24} & {\footnotesize{}0.865} & {\footnotesize{}118.80} & {\footnotesize{}0.200}\tabularnewline
{\footnotesize{}Ours} & {\footnotesize{}\textbf{21.64}} & {\footnotesize{}\textbf{0.868}} & {\footnotesize{}\textbf{113.98}} & {\footnotesize{}\textbf{0.171}}\tabularnewline
\hline
\end{tabular}{\small\par}
\end{table}

\section{Experiments with Center Square Masks}
\begin{figure}
\begin{centering}
\includegraphics[width=0.95\linewidth]{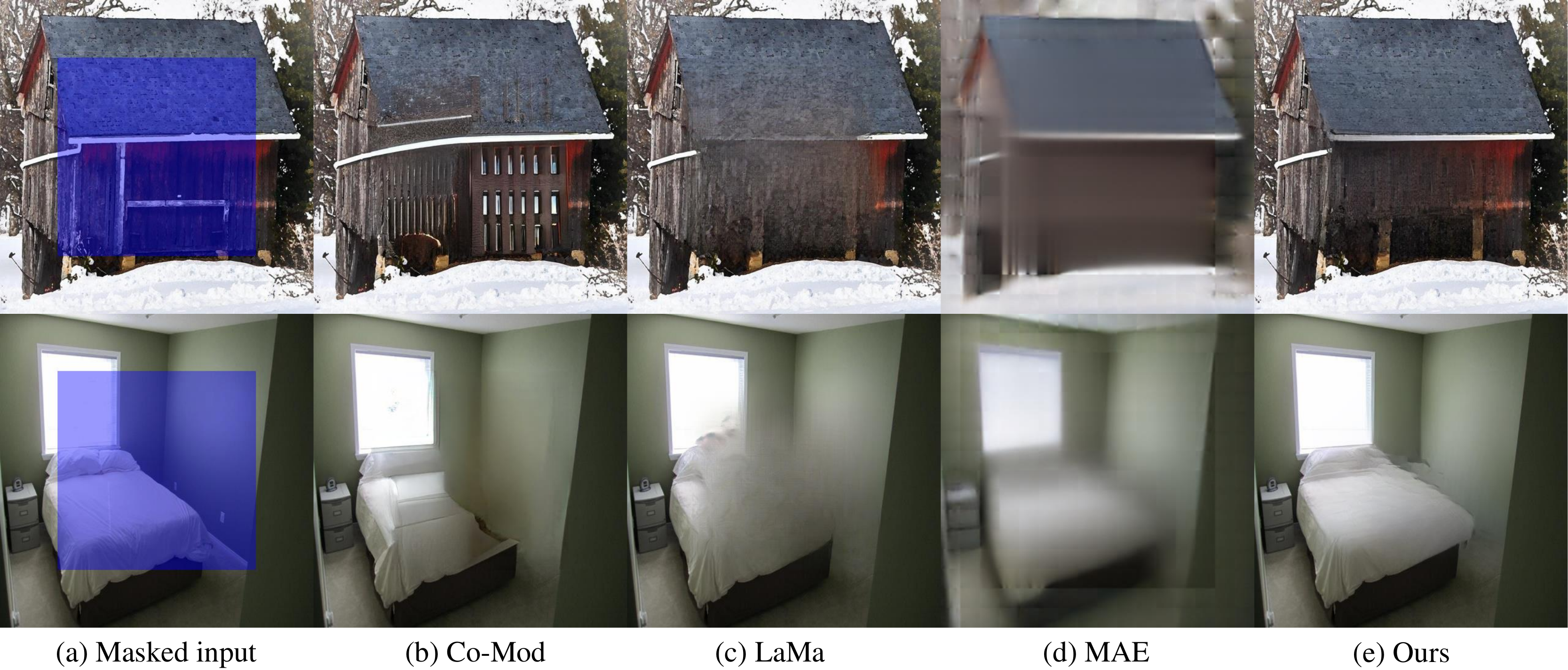}
\par\end{centering}
\caption{Qualitative results with 40\% center square masks on 512$\times$512 Places2.}
\label{fig:square_mask}
\end{figure}

Both quantitative and qualitative results of 512$\times$512 Places2 test set with 40\% center square masks are shown in Tab.~\ref{tab:square_mask} and Fig.~\ref{fig:square_mask} respectively. Note that our method is trained \textit{without any rectangular mask}, while masks of Co-Mod include rectangular ones. Co-Mod suffers from hallucinated artifacts and LaMa tends to generate blur results. Our model beat others with best PSNR and LPIPS even without training on rectangular masks, benefited by MAE priors.

\begin{table}[h]
\small
\centering
\caption{Quantitative results with 40\% center square masks on 512$\times$512 Places2.\label{tab:square_mask}}
\vspace{-0.1in}
\begin{tabular}{c|cccc}
\hline
 & {\footnotesize{}PSNR$\uparrow$} & {\footnotesize{}SSIM$\uparrow$} & {\footnotesize{}FID$\downarrow$} & {\footnotesize{}LPIPS$\downarrow$}\tabularnewline
\hline
{\footnotesize{}Co-Mod} & {\footnotesize{}17.59} & {\footnotesize{}0.755} & {\footnotesize{}\textbf{52.38}} & {\footnotesize{}0.262}\tabularnewline
{\footnotesize{}LaMa} & {\footnotesize{}19.69} & {\footnotesize{}0.801} & {\footnotesize{}61.67} & {\footnotesize{}0.268}\tabularnewline
{\footnotesize{}Ours} & {\footnotesize{}\textbf{19.82}} & {\footnotesize{}\textbf{0.804}} & {\footnotesize{}53.61} & {\footnotesize{}\textbf{0.214}}\tabularnewline
\hline
\end{tabular}
{\small\par}
\vspace{-0.1in}
\end{table}

\section{Ablations about Prior Attention from Different Layers}

In Tab.~\ref{tab:attention_diff_layer}, we test prior attentions of different layers from the start of the MAE decoder, and find that half-layer (4) just enjoys marginally better FID compared with all-layer (8) used in the main paper.
These results show that using such attention priors from MAE is effective, while in general there is no significant difference in using attention from which layer. 

\begin{table}[h] \small 
\centering
\caption{Quantitative results of prior attention layers used from the start of MAE on the Places2 subset.\label{tab:attention_diff_layer}}
{\footnotesize{}}%
\begin{tabular}{c|cccc}
\hline 
{\footnotesize{}Attn layer} & {\footnotesize{}PSNR$\uparrow$} & {\footnotesize{}SSIM$\uparrow$} & {\footnotesize{}FID$\downarrow$} & {\footnotesize{}LPIPS$\downarrow$}\tabularnewline
\hline 
{\footnotesize{}-} & {\footnotesize{}24.34} & {\footnotesize{}0.860} & {\footnotesize{}26.84} & {\footnotesize{}0.117}\tabularnewline
{\footnotesize{}2} & {\footnotesize{}24.50} & {\footnotesize{}0.863} & {\footnotesize{}25.61} & \textbf{\footnotesize{}0.112}\tabularnewline
{\footnotesize{}4} & {\footnotesize{}24.51} & {\footnotesize{}0.862} & \textbf{\footnotesize{}25.38} & {\footnotesize{}0.113}\tabularnewline
{\footnotesize{}6} & \textbf{\footnotesize{}24.54} & {\footnotesize{}0.863} & {\footnotesize{}25.67} & {\footnotesize{}0.114}\tabularnewline
{\footnotesize{}8 (ours)} & {\footnotesize{}24.51} & \textbf{\footnotesize{}0.864} & {\footnotesize{}25.49} & {\footnotesize{}0.113}\tabularnewline
\hline 
\end{tabular}{\footnotesize\par}
\end{table}

\section{Comparing with More SOTA Methods}

We further compare our method with recently proposed ZITS~\cite{dong2022incremental} and MAT~\cite{li2022mat} on Places2 in Tab.~\ref{tab:compare_sota}. Our method can still outperform them with mixed masks.

\begin{table}[h] \small 
\centering
\caption{Quantitative results compared with ZITS~\cite{dong2022incremental} and MAT~\cite{li2022mat} on Places2 with mixed masks.\label{tab:compare_sota}}
\begin{tabular}{c|cccc|cccc}
\hline 
 & \multicolumn{4}{c|}{256$\times$256} & \multicolumn{4}{c}{512$\times$512}\tabularnewline
\hline 
 & PSNR$\uparrow$ & SSIM$\uparrow$ & FID$\downarrow$ & LPIPS$\downarrow$ & PSNR$\uparrow$ & SSIM$\uparrow$ & FID$\downarrow$ & LPIPS$\downarrow$\tabularnewline
\hline 
MAT & 22.37 & 0.841 & 1.68 & 0.134 & 21.68 & 0.838 & 32.43 & 0.165\tabularnewline
ZITS & 24.42 & 0.870 & 1.47 & 0.108 & 24.23 & \textbf{0.881} & 26.08 & 0.133\tabularnewline
Ours & \textbf{24.53} & \textbf{0.871} & \textbf{1.31} & \textbf{0.101} & \textbf{24.33} & 0.880 & \textbf{25.39} & \textbf{0.119}\tabularnewline
\hline 
\end{tabular}
\end{table}

\section{More Qualitative Results}

\begin{figure}
\begin{centering}
\includegraphics[width=0.95\linewidth]{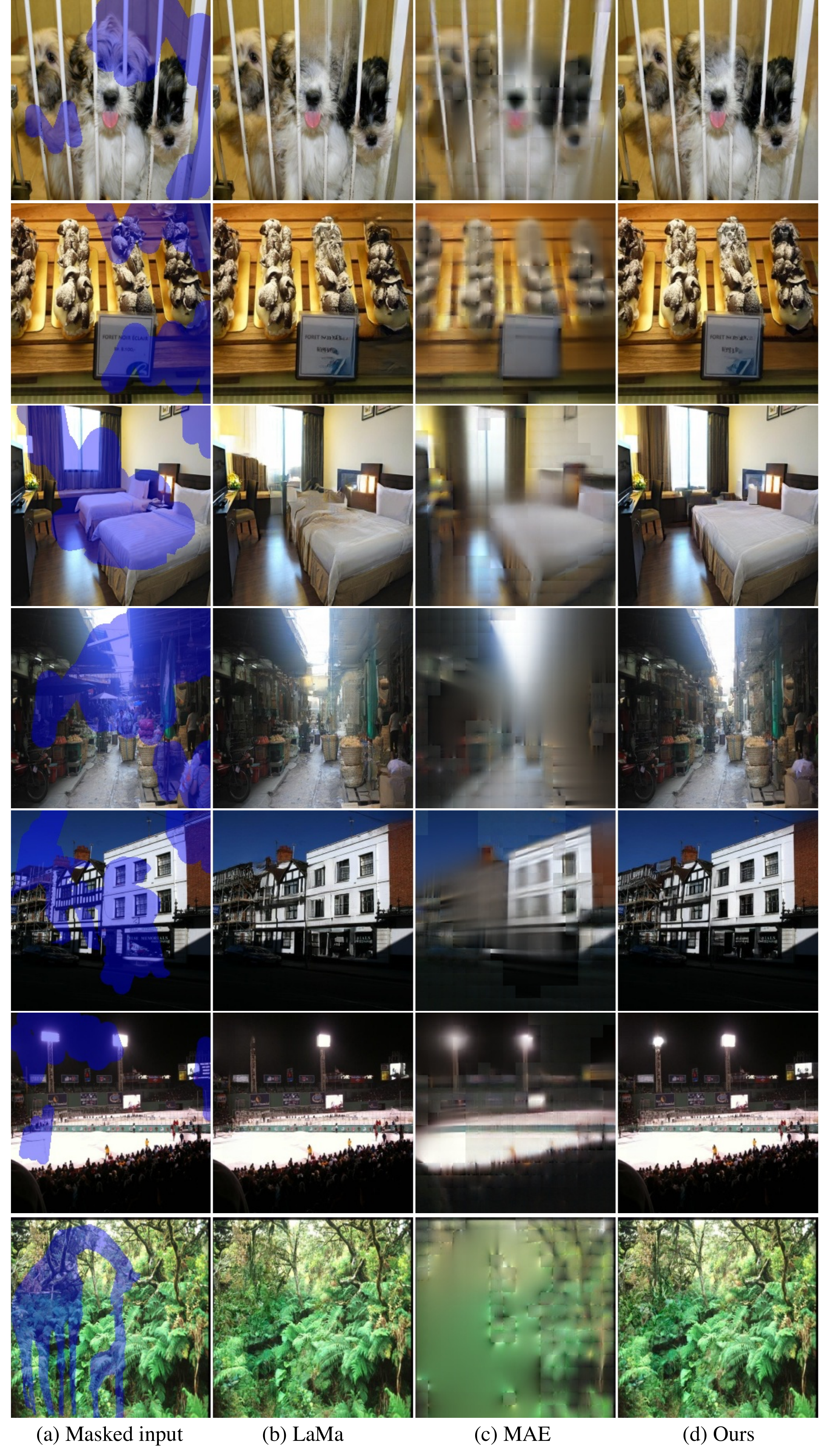}
\par\end{centering}
\caption{Qualitative results of places2 256$\times$256 images. From left to right are masked inputs,  LaMa~\cite{suvorov2021resolution}, MAE~\cite{He2021MaskedAA}, and our results. Please zoom-in for details.}
\label{fig:supp_qualitative_places2_256}
\end{figure}

\begin{figure}
\begin{centering}
\includegraphics[width=0.95\linewidth]{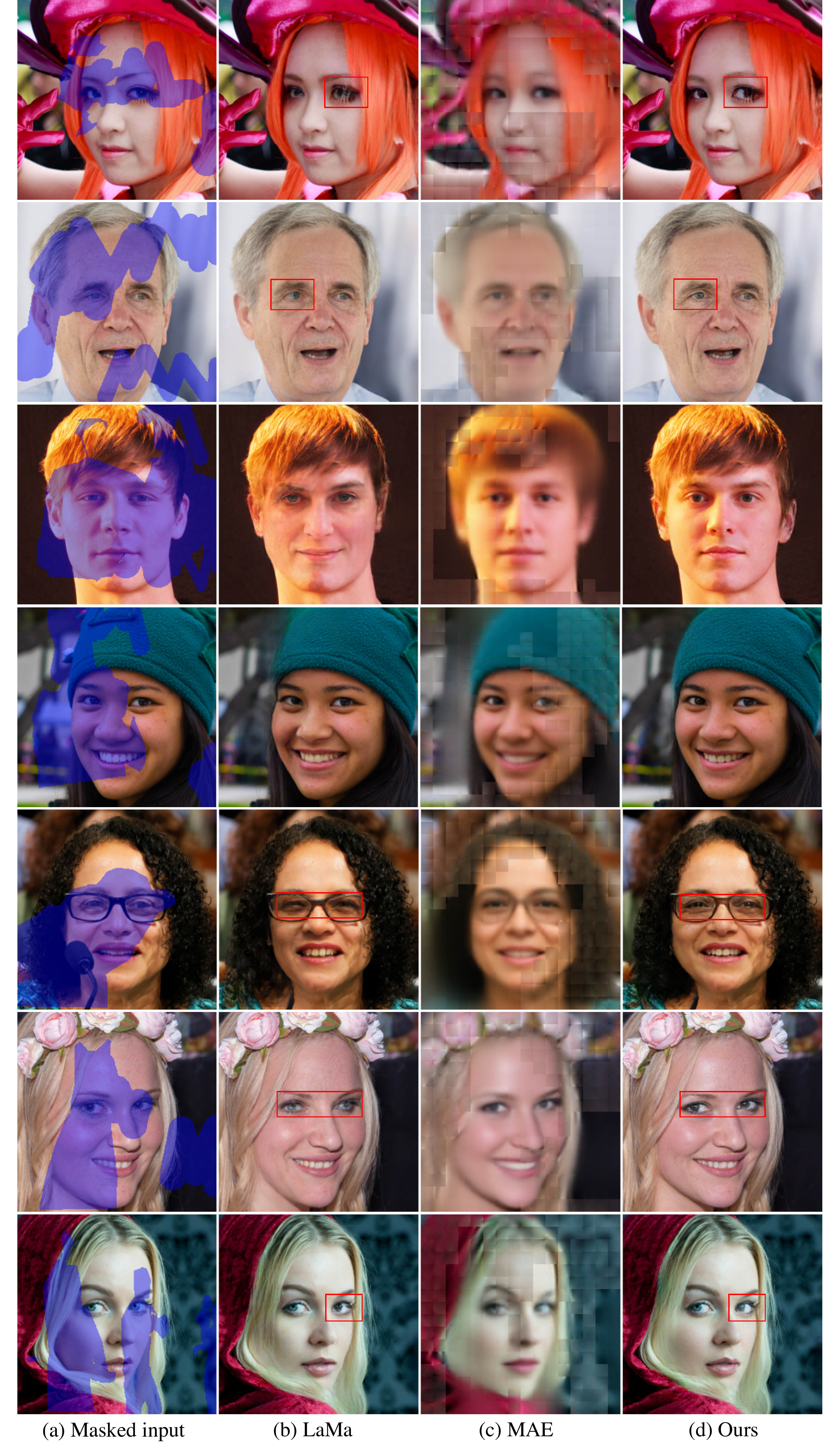}
\par\end{centering}
\caption{Qualitative results of places2 256$\times$256 images. From left to right are masked inputs,  LaMa~\cite{suvorov2021resolution}, MAE~\cite{He2021MaskedAA}, and our results. Please zoom-in for details.}
\label{fig:supp_qualitative_ffhq_256}
\end{figure}

\begin{figure}
\begin{centering}
\includegraphics[width=0.95\linewidth]{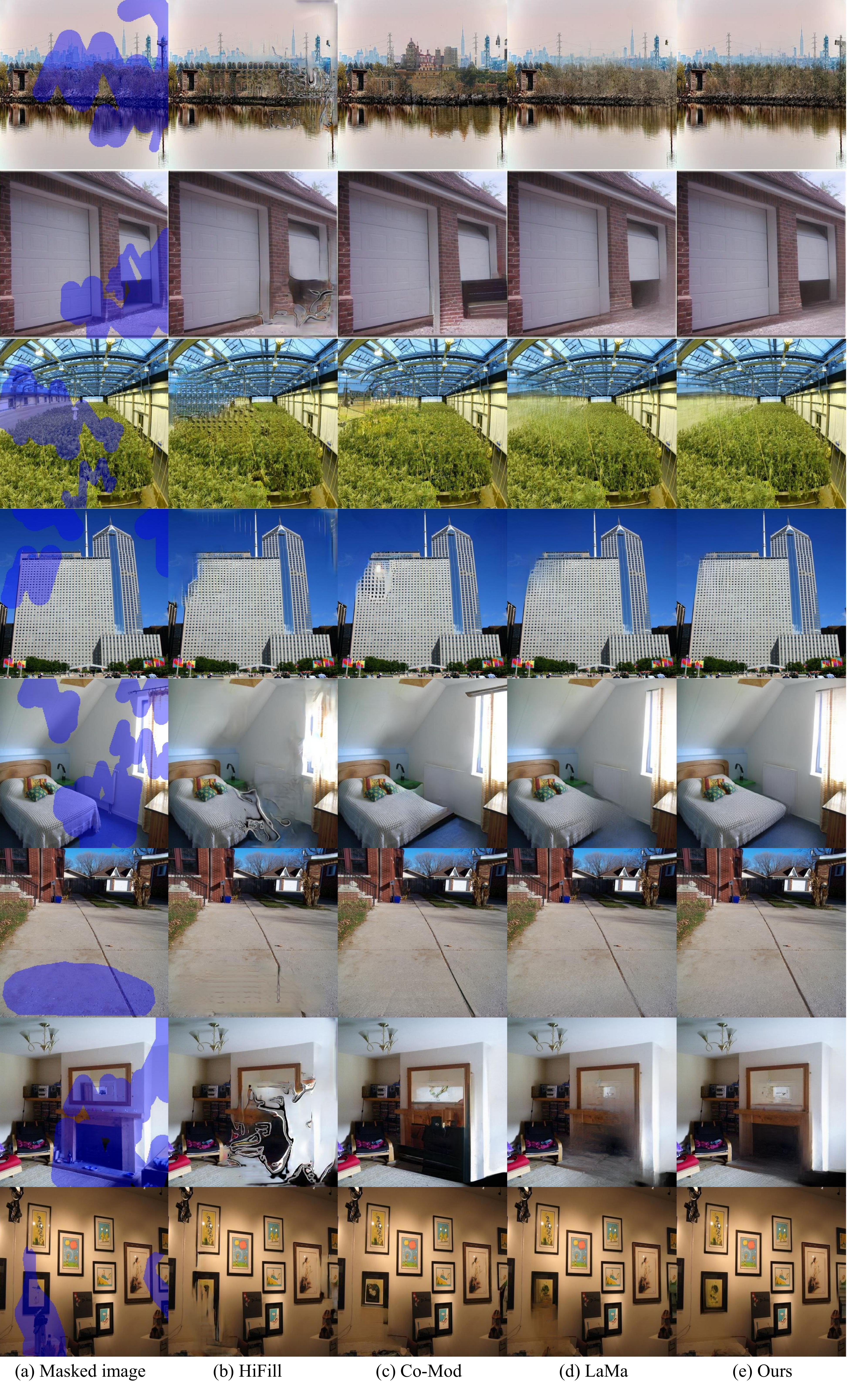}
\par\end{centering}
\caption{Qualitative results of 512$\times$512 images from Places2. From left to right are masked image, HiFill~\cite{yi2020contextual}, Co-Mod~\cite{zhao2021large},  LaMa~\cite{suvorov2021resolution}, and our results. Please zoom-in for details.}
\label{fig:supp_512}
\end{figure}

\begin{figure}
\begin{centering}
\includegraphics[width=0.95\linewidth]{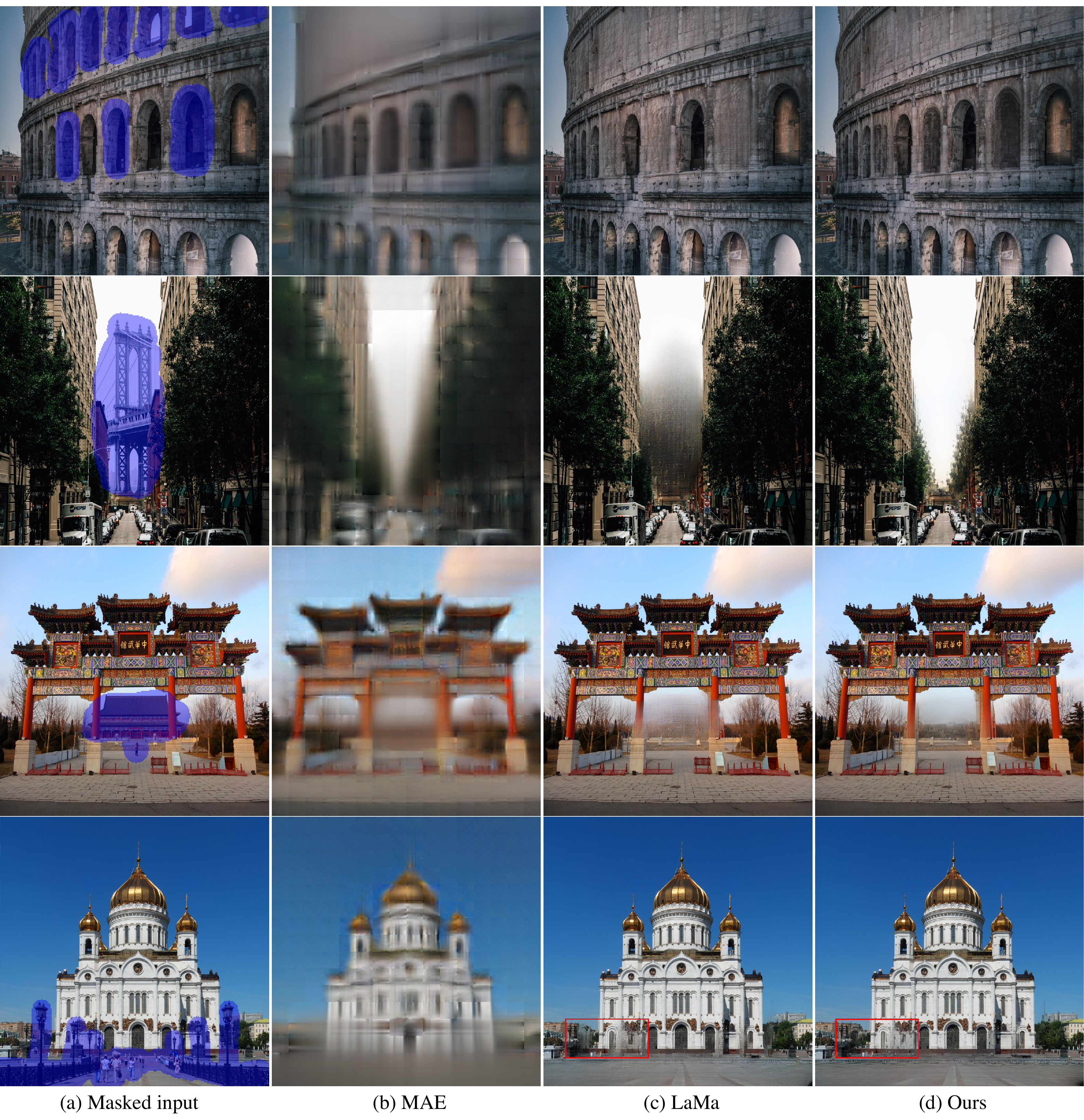}
\par\end{centering}
\caption{Qualitative results of 1024$\times$1024 images from DIV2K. From left to right are masked inputs, MAE~\cite{He2021MaskedAA}, LaMa~\cite{suvorov2021resolution}, and our results. Please zoom-in for details. For the first picture, both LaMa and our method fill all holes in the first row of the Colosseum, but our method still remains the holes in the second row. Because the MAE result is learned with a global receptive field, which guides our method to inpaint a more reasonable result rather than copying meaningless textures nearby.}
\label{fig:supp_1k_res}
\end{figure}

\begin{figure}
\begin{centering}
\includegraphics[width=0.95\linewidth]{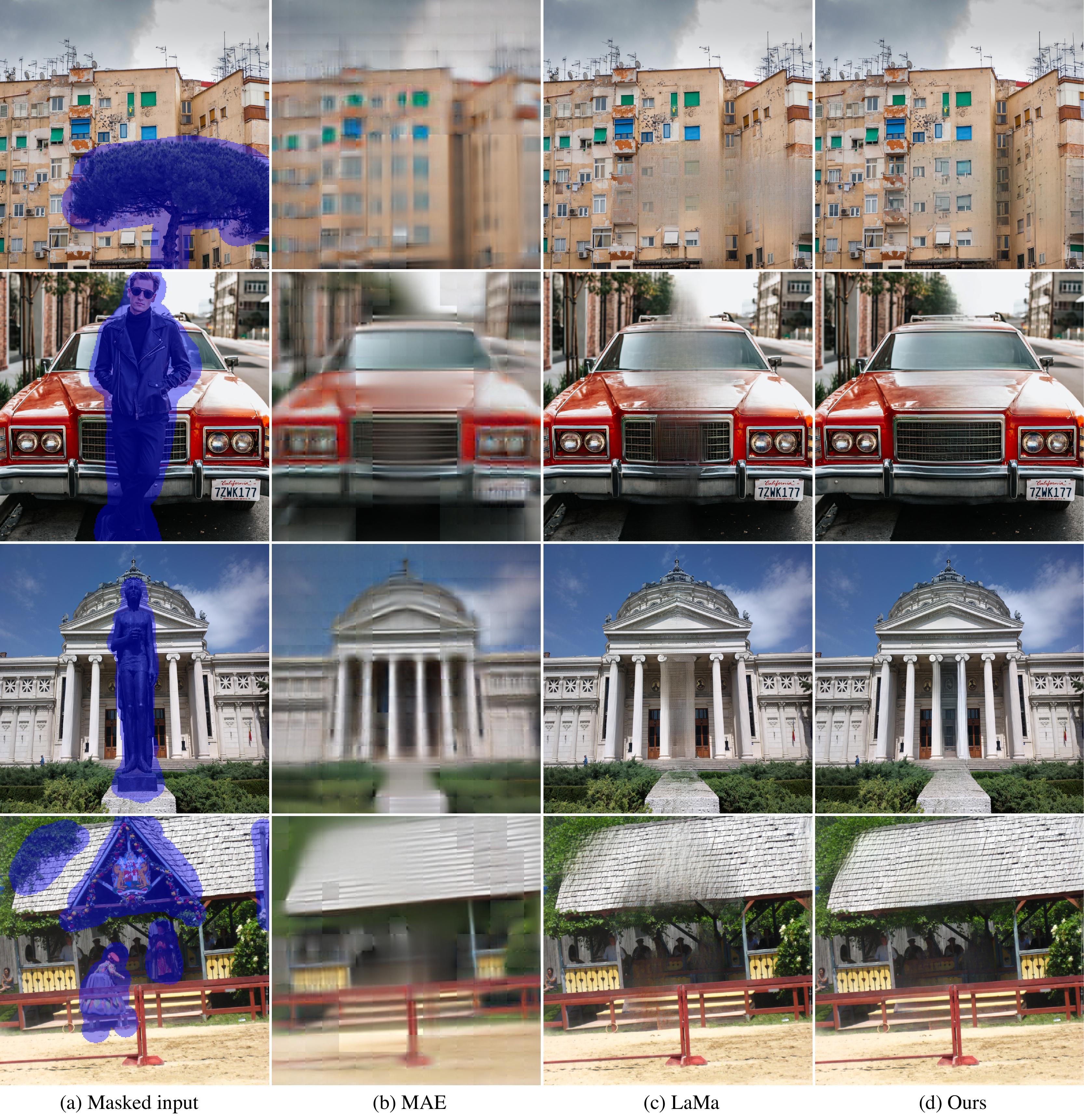}
\par\end{centering}
\caption{Qualitative results of 1024$\times$1024 images, which have also been shown in the main paper. From left to right are masked inputs, MAE~\cite{He2021MaskedAA}, LaMa~\cite{suvorov2021resolution}, and our results. Please zoom-in for details.}
\label{fig:supp_1k_res_2}
\end{figure}

More 256$\times$256 results of Places2 and FFHQ are shown in Fig.~\ref{fig:supp_qualitative_places2_256} and Fig.~\ref{fig:supp_qualitative_ffhq_256} respectively. For face images, we recommend to zoom-in for details near the eye regions. Our method tends to generate consistent eyes for face inpainting. We also provide more 512$\times$512 results of Places2 in Fig.~\ref{fig:supp_512}, and some 1024$\times$1024 results from DIV2K~\cite{Agustsson_2017_CVPR_Workshops} in Fig.~\ref{fig:supp_1k_res}. For the HR inpainting, we find an interesting phenomenon that the MAE enhanced results enjoy larger receptive fields for the structural recovery in HR cases as shown in the first row of Fig.~\ref{fig:supp_1k_res}. Besides, for a better reading experience, 1k results shown in the main paper are slightly compressed. We show the high quality ones in Fig.~\ref{fig:supp_1k_res_2}.

\section{Limitations and Future Works}
\begin{figure}
\begin{centering}
\includegraphics[width=0.7\linewidth]{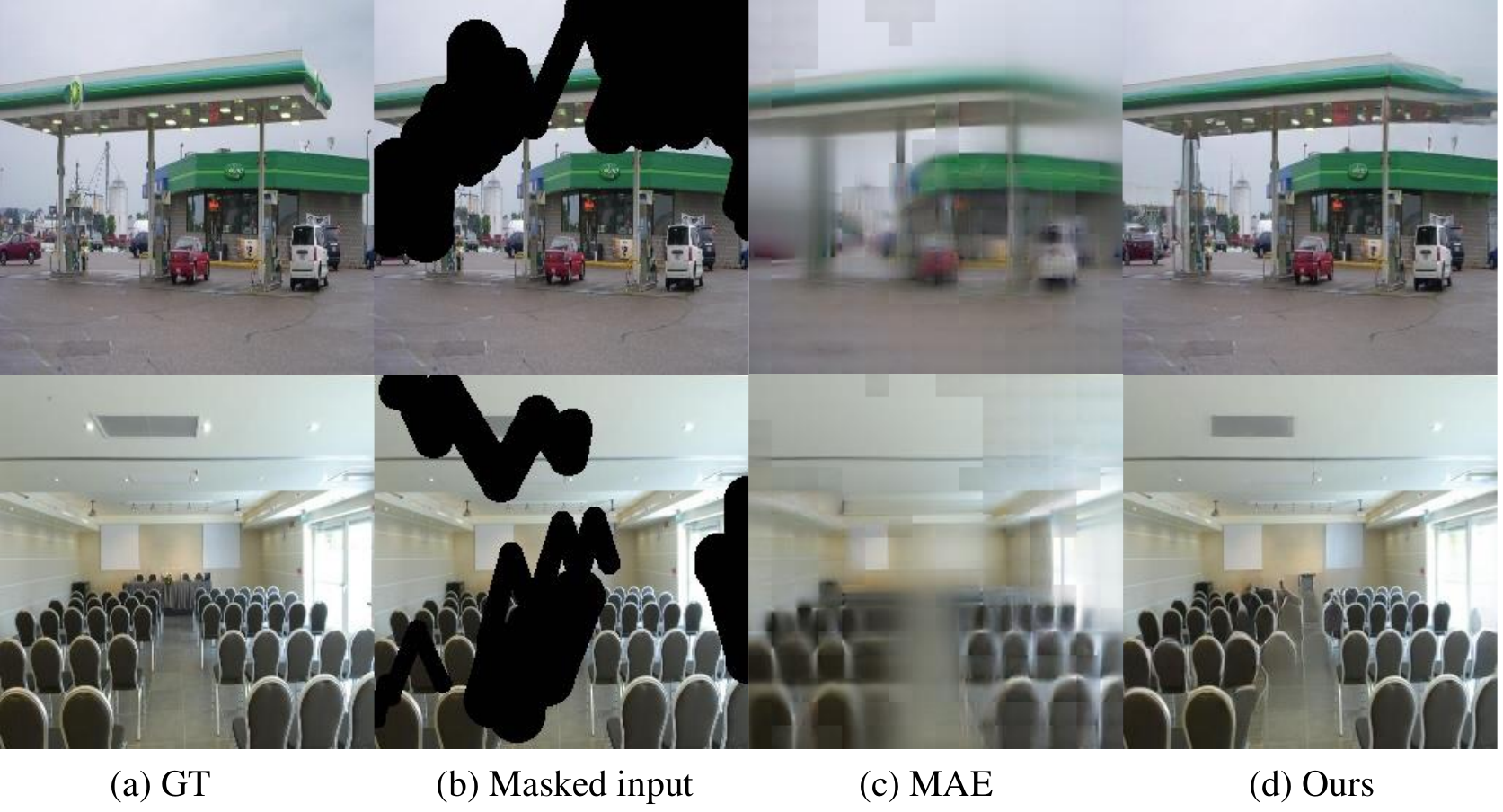}
\par\end{centering}
\caption{Failed cases of our method. GT indicates ground truth images}
\label{fig:fail}
\end{figure}

Although our proposed FAR is powerful enough to inpaint impressive results, it still suffers from fail cases as shown in Fig.~\ref{fig:fail}. MAE has some difficulty in exactly recovering the object/building boundaries or some complex man-made structures, which leads to some ambiguity. To tackle this problem, we think that structure priors can provide more exact boundaries for high-fidelity results. Besides, as mentioned in the main paper, an interesting future work would be exploring features from different MAE layers for inpainting. In our opinion, such improvements are orthogonal to other proposed components in this paper. Our pre-trained Places2 MAE will be released, which is benefit for the community to further study the representation learning based image inpainting. 
Moreover, although our MAE pre-trained on Places2 is generalized enough for the inpainting, pre-training MAEs on larger datasets (such as ImageNet-22K~\cite{ILSVRC15} or even JFT-3B~\cite{zhai2022scaling}) may achieve superior downstream performance.

\clearpage
\bibliographystyle{splncs04}
\bibliography{egbib}
\end{document}